\documentclass[letterpaper, 10pt, journal, twoside]{IEEEtran}  % RA-L camera-ready format

\IEEEoverridecommandlockouts                              % This command 

\usepackage{subcaption}
\usepackage{capt-of}
\usepackage{graphics} % for pdf, bitmapped graphics files
\usepackage{epsfig} % for postscript graphics files
\usepackage{array} % for extended column formatting
\usepackage[export]{adjustbox} % for image alignment in tables
\usepackage{newtxtext} % Times text font (modern mathptmx/times replacement)
\usepackage{amsmath} % assumes amsmath package installed
\usepackage{newtxmath} % Times math font; also provides amssymb/amsfonts symbols
\usepackage[utf8]{inputenc} % allow utf-8 input
\usepackage[T1]{fontenc}    % use 8-bit T1 fonts
\usepackage{hyperref}       % hyperlinks
\hypersetup{hidelinks}      % no colored boxes around links/citations; comment out to see links while reviewing
\usepackage{url}            % simple URL typesetting
\usepackage{booktabs}       % professional-quality tables
\usepackage{nicefrac}       % compact symbols for 1/2, etc.
\usepackage{microtype}      % microtypography
\usepackage{balance}        % balance columns on the last page (\balance before bibliography)
\usepackage{siunitx}        % consistent number/unit typesetting
\DeclareSIUnit\fps{fps}
\usepackage[dvipsnames]{xcolor}         % colors
\usepackage{dirtytalk}
\usepackage{multirow}
\usepackage{dblfloatfix}
\usepackage{float}
\usepackage{tikz}
\usepackage{makecell}

\usepackage{pifont}
\usepackage[normalem]{ulem} % \sout strikethrough for marked deletions; normalem keeps \emph italic
\newcommand{\greentick}{\textcolor{ForestGreen}{\checkmark}}
\newcommand{\redx}{\textcolor{red}{\text{\sffamily \ding{55}}}}
\usepackage[numbers,sort]{natbib}
\title{GHOST in the Robots:\\Real-Time Exocentric Dual-Robot VR Teleoperation from Onboard Cameras}

\author{Yichen Wei$^*$, Faisal Zaghloul$^*$, Soujanya C~Aryal, Aanya K.~Agrawal, Chengfan Li, Jason Xinyu Liu$^{\ddagger}$,\\James Tompkin, and Stefanie Tellex% <-this % stops a space
\thanks{Manuscript received: March 20, 2026; Revised June 25, 2026; Accepted August 2, 2026.}%
\thanks{This paper was recommended for publication by Editor Ki-Uk Kyung upon evaluation of the Associate Editor and Reviewers' comments.
This work was supported by NASA 80NSSC23M0075, NSF CNS-2038897, and Amazon.}%
\thanks{$^{*}$Equal contribution.}%
\thanks{Yichen Wei, Faisal Zaghloul, Soujanya C~Aryal, Aanya K.~Agrawal, Chengfan Li, James Tompkin, and Stefanie Tellex are with the Department of Computer Science, Brown University, Providence, RI 02912, USA {\tt\footnotesize stefanie\_tellex@brown.edu}.}%
\thanks{$^{\ddagger}$Jason Xinyu Liu is with CSAIL, Massachusetts Institute of Technology, Cambridge, MA 02139, USA. Work completed while at Brown University.}
\thanks{Digital Object Identifier (DOI): 10.1109/LRA.2026.3726328.}%
}

\makeatletter
\def\@IEEEpubidpullup{0pt}
\makeatother
\IEEEpubid{%
  \raisebox{\dimexpr-\footskip-1em\relax}[0pt][0pt]{%
    \parbox[b]{\textwidth}{%
      \centering\scriptsize
      \copyright~2026 IEEE. Personal use of this material is permitted. Permission from IEEE must be obtained for all other uses, in any current or future media, including reprinting/republishing this material for advertising or promotional purposes, creating new collective works, for resale or redistribution to servers or lists, or reuse of any copyrighted component of this work in other works.%
    }%
  }%
}

\makeatletter
\let\@oldmaketitle\@maketitle
\renewcommand{\@maketitle}{%
  \@oldmaketitle
  \vspace{.5em}%

  \begin{@twocolumnfalse}
    \def\@captype{figure}%
    \setcounter{figure}{0}%
    \refstepcounter{figure}% Set up figure 1 properly
    \begin{minipage}{\textwidth}
      \centering
    \includegraphics[width=0.95\linewidth]{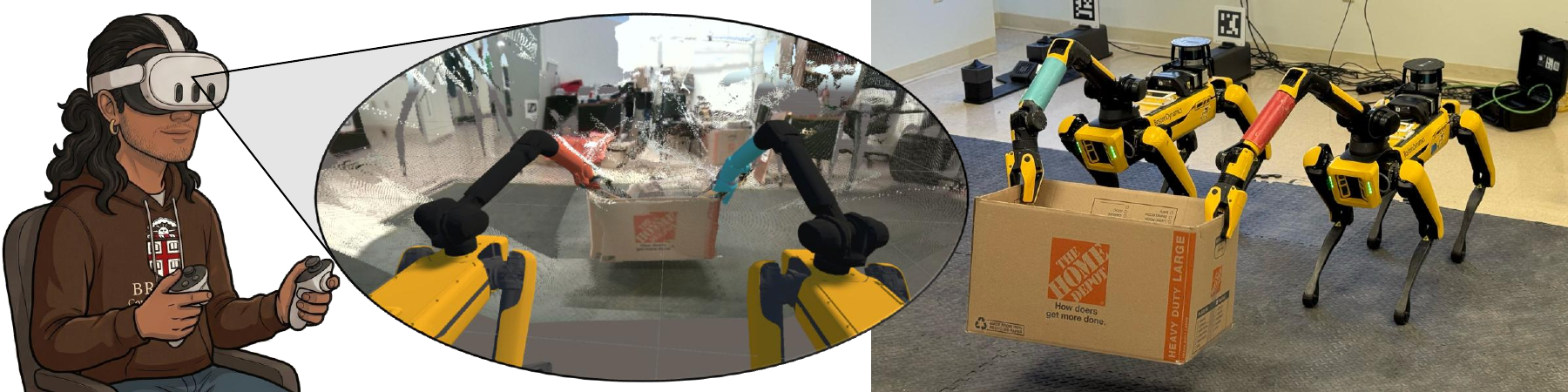}
      \addtocounter{figure}{-1}% Go back to 0 temporarily
    \captionof{figure}{
      \textbf{GHOST enables single-operator dual-robot control.}
      Our VR teleoperation system enables individual or simultaneous dual-robot control to perform complex mobile manipulation tasks, with a 3D virtual scene reconstructed and rendered in real time from onboard cameras.
    }
    \label{fig:main_splash}%
    \end{minipage}
  \end{@twocolumnfalse}

  \vspace{-1em}%
}
\makeatother

\begin{document}

\bstctlcite{BSTcontrol} % force `et al.' for >6 authors in the bibliography

\maketitle

%%%%%%%%%%%%%%%%%%%%%%%%%%%%%%%%%%%%%%%%%%%%%%%%%%%%%%%%%%%%%%%%%%%%%%%%%%%%%%%%
\begin{abstract}
Teleoperating multiple robots simultaneously enables additional views and coordinated control.
Yet, it poses fundamental challenges: the system must present sensor data cohesively and allow operators to manage multiple robot bases, arms, and cameras while maintaining low latency.
Current multi-robot teleoperation systems require multiple operators, rely on autonomy, or restrict operators to high-level commands.
We present GHOST: an open-source VR teleoperation system that enables single-operator control of two mobile manipulators via direct low-level commands using only onboard sensing.
GHOST creates an exocentric 3D workspace by aligning real-time point clouds from the robots' RGB-D cameras, where scene coverage is improved through learning-based completion to aid operator spatial awareness.
For control, the operator uses a mode-switching architecture to command either robot individually or both robots simultaneously.
Experiments with 15 novice participants demonstrate 1.6--4$\times$ the success rate of an off-the-shelf tablet interface.
For experts across nine challenging dual-robot tasks, our system enabled completion of two tasks that were infeasible with the tablet, and was 1.47$\times$ faster on average than the tablet.
Website and code: \url{https://h2r.github.io/GHOST/}.
\end{abstract}


\IEEEpeerreviewmaketitle

%%%%%%%%%%%%%%%%%%%%%%%%%%%%%%%%%%%%%%%%%%%%%%%%%%%%%%%%%%%%%%%%%%%%%%%%%%%%%%%%

\vspace{1cm}
\section{Introduction}
\vspace{-0.07cm}

\IEEEPARstart{T}{eleoperation} systems aim to enable remote robot control in complex unstructured environments. This is often attempted via virtual reality (VR) because immersing operators in a virtual space with the robot can provide stereo depth perception, improved spatial awareness, and precise, intuitive control.
Existing VR interfaces typically control only one robot.  However, real-world tasks like transporting large or elongated objects require multiple support points, and this could be achieved with multiple robots. 
In addition, single-robot control often constrains the operator's field of view due to limited camera angles and arm occlusions that create blind spots. This forces operators to work with incomplete information or frequently reposition the robot to adjust their view.

Deploying multiple robots in VR teleoperation introduces new challenges to single-operator perception and control.
The perception system must present precise and coherent spatial information to the operator.
Additionally, operators must divide their limited attention across multiple robot bases and arms~\cite{wickens2015stom_task_switching_model, cowan2010magical,wickens2008multiple,tung2021collab_teleop}.
Thus, the control system must efficiently map the operator's inputs to the high-dimensional robot control space.
Lastly, the system overall must provide responsive control with multiple video streams, 3D scene data, and synchronized commands, requiring careful system design to maintain low latency over wireless connections.
Existing multi-robot systems typically cannot meet these challenges. They may rely upon multiple operators~\cite{tung2021collab_teleop}, autonomy with only occasional human intervention~\cite{swamy2020scaled_autonomy}, or constrain operators to only high-level commands rather than precise low-level control~\cite{roldan2017multi, engelbracht2025spot_on}.

To address these challenges, we present
GHOST (GHOST Human Operator for Simultaneous Teleoperation), a system that enables a single operator to control two mobile manipulators, individually or simultaneously, from within an immersive virtual scene, as if the operator were an invisible ghost standing alongside the robots.
Our system reconstructs a 3D point cloud by processing RGB-D streams from the robots' onboard cameras.
As each camera only provides sparse depth information, GHOST applies real-time depth completion to infer missing regions, then aligns the completed point clouds into a coherent virtual scene. This expands the operator's field of view and reduces the blind spot problem.
The control scheme allows manipulation and navigation, enabling intuitive and real-time 6-DoF pose control of two mobile manipulators through two hand controllers.
Simultaneous control of both robots through only two hand controllers is possible because their relative base poses can be `locked', keeping cognitive load low.
Our system design that combines real-time scene reconstruction, an immersive exocentric view, and low-level multi-robot control has not been previously shown.

We evaluate the system on nine teleoperation tasks that are difficult or impractical with a single robot but simpler with two. We compare to an off-the-shelf tablet interface.
Across both novice and expert operators, our system improved task success rate and typically also completion time.
For 15 novices, who attempted two easier tasks, our system improved the success rate by 1.6--4$\times$.
For three experts, our system enabled completion of two tasks that were infeasible with the tablet, and the remaining seven tasks were completed 1.47$\times$ faster on average.
An ablation replacing the 3D scene with RGB camera feeds showed that, for experts, direct 6-DoF control provides much of the speed benefit, while the reconstructed 3D scene improves task completion reliability.
Taken together, these results show that our new VR-based teleoperation system can provide improvements in efficiency and reliability and also enable new multi-robot capabilities on existing hardware.

We make the following contributions: 
(1) An exocentric 3D scene for multi-robot teleoperation;
(2) A low-latency depth completion pipeline; 
(3) A mode-switching control design for dual-robot teleoperation; and
(4) A set of challenging evaluation tasks for dual-robot teleoperation.

\begin{table*}[ht!]
    \footnotesize
    \centering
    \caption{\textbf{GHOST is a new point in the multi-robot teleoperation design space.} Existing multi-robot teleoperation works did not produce real-time 3D scene depictions with low-level multi-robot control.}
    \begin{tabular}{l  c c c l c c c}
    \toprule
    Work                                                           & \makecell{Novel\\environments?} & \makecell{Single\\operator?} & \makecell{Low-level\\actions?} & \makecell{Multi-robot coordination} & \makecell{Real-time\\scene?} & \makecell{3D\\scene?} & \makecell{Onboard\\cameras?}       \\
    \midrule
    Lewis et al.\ \cite{lewis2011two}                              & \greentick                    & \greentick                 & \greentick                   & Mirror                              & \greentick                 & \redx               & \greentick                \\
    Roldan et al.\ \cite{roldan2017multi}                           & \redx                         & \greentick                 & \redx                        & High-level action only              & \redx                      & \greentick          & \redx                     \\
    CollabTeleop \cite{tung2021collab_teleop}                     & \greentick                    & \redx                      & \greentick                   & N/A                                 & \greentick                 & \redx               & \redx                \\
    Body Extension \cite{hirao2023body_extension_two_mobile_manip} & \greentick                    & \greentick                 & \greentick                   & Individual control                  & \greentick                 & \redx               & \greentick                \\
    Laghi et al.\ \cite{laghi2018shared}                           & \redx                         & \greentick                 & \greentick                   & Pivot                               & \redx                      & \redx               & \redx                     \\
    Ozdamar et al.\ \cite{ozdamar2022shared}                       & \redx                         & \greentick                 & \greentick                   & Pivot                               & \redx                      & \redx               & \redx                     \\
    Zick et al.\ \cite{zick2024teleoperation}                       & \redx                         & \greentick                 & \redx                        & High-level action only              & \greentick                 & \redx               & \redx                     \\
    Scaled Autonomy \cite{swamy2020scaled_autonomy}                    & \greentick                    & \greentick                 & \greentick                  & Autonomous + occasional individual control                 & \redx                      & \redx               & \redx                     \\
    Spot-On \cite{engelbracht2025spot_on}                          & \redx                         & \greentick                 & \redx                        & High-level action only              & Partial                    & \greentick          & \redx                     \\
    Chen et al.\ \cite{chen2023mr_teaming}                        & \greentick                    & \greentick                 & \redx                        & High-level action only              & \greentick                 & \greentick          & \greentick                \\
    Ours                                                          & \greentick                    & \greentick                 & \greentick                   & \textbf{Pivot + individual control} & \greentick                 & \greentick          & \greentick                \\ 
    \bottomrule
    \end{tabular}
    \label{tab:related}
    %\vspace{-2em}
\end{table*}

\section{Related Work}

\subsection{Visual Representations for Teleoperation}

Robot teleoperation interfaces generally follow two representational paradigms: egocentric and exocentric \cite{lipton2017baxter,whitney2018ros_reality}. \textbf{Egocentric} interfaces, common in VR teleoperation \cite{cheng2025opentv, chuang2025active}, present the environment from the robot's perspective and map commands to its frame. However, camera motion delays relative to human head and eye motion can induce motion sickness \cite{xiong2025vision_in_action}. In multi-robot settings, operators need to mentally reconstruct spatial relationships by switching between cameras, increasing cognitive load. Simultaneous egocentric control is also impractical, as it requires separate live inputs, e.g., two tablet interfaces. GHOST instead uses an exocentric design that unifies multi-robot coordination in a shared 3D space.

\textbf{Exocentric} interfaces visualize robots and sensor data in a global frame, providing third-person views that naturally support multi-robot coordination \cite{bauer2024ghost}. Existing exocentric VR systems may require external capture cameras \cite{engelbracht2025spot_on}, limiting operational scope. Additionally, accurate 3D reconstruction remains challenging: stereo depth is sparse; Gaussian splatting and neural radiance fields \cite{10801345} produce high-quality renderings but are too slow for real-time reconstruction. This motivates our use of depth completion models.

\subsection{Depth Completion}

One challenge that arises when generating point clouds is the sparsity of the input depth data.
Depth completion techniques attempt to fill holes in depth camera data caused by low-reflectance objects, multi-path light effects, or distant scene regions, using an RGB image as a guide.
Earlier methods used convolutional neural networks (CNNs) with different strategies to fuse depth features \cite{cheng2019learning, nconv, ma2018sparse, bpnet}, while more recent works use vision transformers (ViT) \cite{vit, oquab2023dinov2, yang2024depth} and diffusion models \cite{DBLP:journals/corr/abs-2112-10752} to improve depth completion quality \cite{lin2025promptda, zhang2023completionformer, viola2024marigolddc}.

For interactive robotics scenarios, we require a depth completion method that provides high-quality predictions, generalizes well to novel environments, and maintains an interactive frame rate.
For our system, the ViT-based model PromptDA \cite{lin2025promptda} best fits our needs, providing a balance between runtime and quality. 
PromptDA uses an adapter head that processes and fuses features from the transformer backbone at various layers with the input color and depth maps. 
While emerging works can produce point clouds from a batch of RGB(-D) images \cite{wang2025vggt, keetha2025mapanything}, we found the resulting point clouds to be too low-quality for interactive real-time use.

\subsection{Control Systems for Teleoperation}
Beyond visual feedback, effective teleoperation requires control interfaces that balance ease of use and precise robot control. For navigation, direct physical guidance \cite{fu2024mobilealoha} provides intuitive control but is not remote teleoperation. Point-and-click interfaces on maps, images, or the physical scene \cite{zick2024teleoperation,kemp2008point,gu2022ar_point_click} use high-level path planning but lack low-level control. Joysticks can offer both precise and remote control \cite{dass2024telemoma}. Learned controllers can generate base motion automatically while the operator teleoperates only the arm \cite{honerkamp2025moma-teleop}. 

For manipulation, direct physical guidance of the robot \cite{maccio2024kinesthetic} or a handheld gripper \cite{yang2023moma_force} again precludes remote teleoperation. Velocity-based devices, from familiar joysticks to 6-DoF masters and space mice \cite{ryu2010_6dof_spacemouse,dhat2024using3dmice}, are less intuitive for mapping to 6D pose \cite{long2016effect_of_control_device}.
Joint-copy and leader-follower systems enable direct joint control but require platform-specific leader hardware, such as robot-arm leaders \cite{zhao2023aloha} or low-cost printed replicas \cite{wu2024gello}, that limits generalizability.
Hand gesture tracking \cite{cheng2025opentv,sivakumar2022robotic_telekinesis,qin2023anyteleop,qiu2025humanoid,posadas2025beavr} can aid dexterous manipulation and imitation learning data collection if the tracking cameras have a clear view of the hands.
VR systems with controller pose tracking \cite{mandlekar2018roboturk,whitney2018ros_reality,lin2024hato,iyer2025openteach} offer promise as an intuitive manipulation control.

\subsection{Multi-robot Teleoperation}

Multi-robot coordination introduces challenges in information fusion, inter-robot communication, and shared operator attention, and prior systems span several design choices (Tab.~\ref{tab:related}). \textbf{High-level autonomous systems} such as Spot-On \cite{engelbracht2025spot_on} and others \cite{roldan2017multi,zick2024teleoperation,chen2023mr_teaming} address these challenges through semi-autonomous behaviors and interfaces for scene exploration or navigation goals. However, they often rely on pre-built maps \cite{roldan2017multi}, static 3D overlays \cite{engelbracht2025spot_on}, or 2D abstract interfaces \cite{zick2024teleoperation}, assuming known environments and precluding low-level commands. Most similar in spirit, Chen et al.\ \cite{chen2023mr_teaming} co-localize multiple robots and an MR (Mixed Reality) headset using onboard sensors and overlay a live dense 3D reconstruction, but restrict the operator to high-level drag-and-drop goal poses executed by an autonomous path planner, targeting navigation rather than low-level mobile manipulation.

At the opposite end, \textbf{low-level multi-operator systems} such as CollabTeleop \cite{tung2021collab_teleop} provide direct control by assigning one operator per robot, but increase human-resource and coordination costs while limiting situational awareness to single onboard cameras. Between these extremes, \textbf{low-level single-operator} approaches use coordinated control strategies: shared pivot points \cite{laghi2018shared,ozdamar2022shared} control multiple robots as a group, while leader-follower architectures \cite{lewis2011two} make secondary robots mirror a primary robot. Attention-scheduling methods \cite{swamy2020scaled_autonomy} instead learn operator preferences to decide which robot receives control, but assume otherwise autonomous robots needing only occasional intervention. These systems support simultaneous control or attention-switching, but offer limited independent execution \cite{laghi2018shared} and typically provide only RGB streams \cite{lewis2011two,tung2021collab_teleop}, which are insufficient for precise spatial reasoning.

Among these, Body Extension \cite{hirao2023body_extension_two_mobile_manip} is closest to ours, assigning one arm to each robot. However, it lacks both integrated 3D vision and coordinated base control, limiting scene understanding and precluding tasks such as cooperative carrying. No existing work provides a single operator with low-level independent and synchronized control of multiple mobile manipulators while maintaining real-time 3D awareness of a shared workspace.

\section{System Overview}
Our system provides single-operator teleoperation of two Boston Dynamics Spot mobile manipulators within VR.
Each Spot consists of a quadruped base, a 6-DoF arm with a gripper and gripper camera, a LiDAR unit on the back, and five body-mounted RGB-D cameras, of which we use the front two for scene reconstruction. 
The robots communicate via onboard 802.11n Wi-Fi with a ROS~2 \cite{ros2} server and a VR client.

\begin{figure*}[t]
\centering
    \begin{minipage}{1.0\linewidth}
    \includegraphics[width=\linewidth]{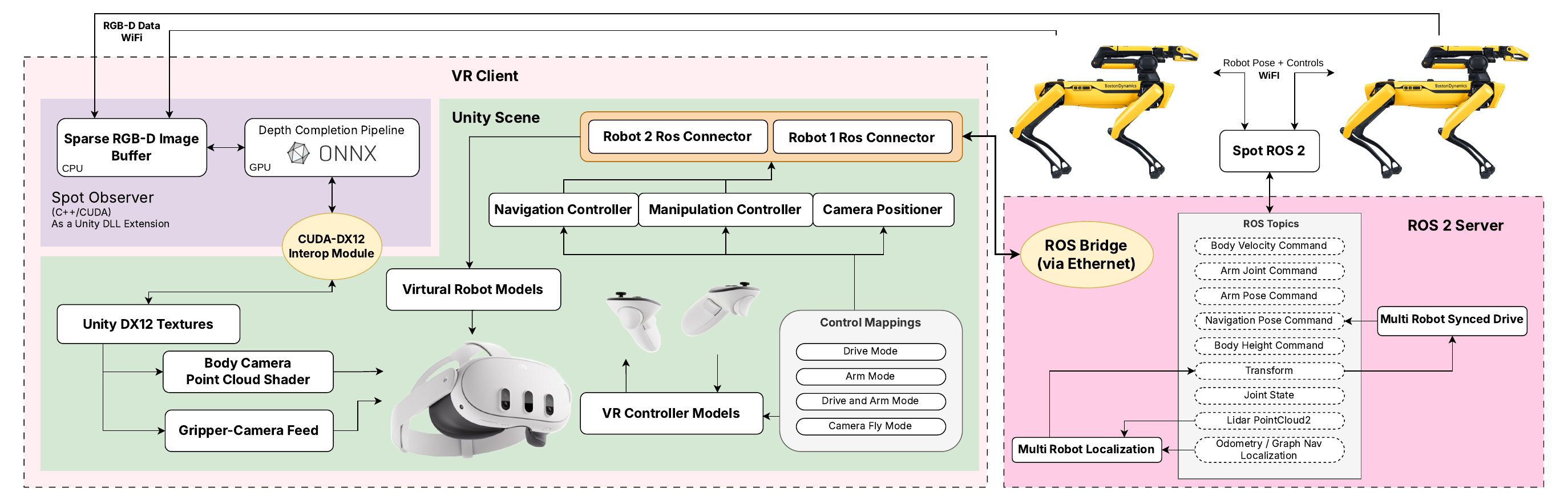}
    \caption{
        \textbf{A diagram of the GHOST system.}
        The system integrates multi-robot ROS~2 communication, a GPU-accelerated RGB-D processing pipeline, and a Unity-based VR interface. Low-bandwidth state/command data are bridged through ROS~2 and drive the virtual robot models, while high-bandwidth RGB-D streams are processed on the GPU and rendered as interactive point clouds for low-latency teleoperation.
    }
    \label{fig:system_diagram}
    \end{minipage}
    %\vspace{-0.25cm}
\end{figure*}
 
The system integrates off-the-shelf tools (PromptDA, ROS~2, Unity, Spot SDK) into three components (Fig.~\ref{fig:system_diagram}):
(1)~A \textbf{Unity-based VR client} that renders the visual feedback and provides the operator with mode-switching control of either robot individually or both simultaneously;
(2)~A \textbf{ROS~2 server} that manages low-bandwidth state streaming (i.e., small message data), command execution, multi-robot localization, and synchronized navigation;
(3)~\textbf{SpotObserver}, a custom GPU-accelerated library on the VR client that bypasses ROS~2 messaging to retrieve RGB-D data directly from each robot, performs real-time depth completion, and produces dense point clouds with minimal latency for Unity to render.
The next two sections describe what these components provide to the operator: visual feedback (Sec.~\ref{sec:vision_sys}) and control (Sec.~\ref{sec:control_design}).

\section{Visual Feedback}
\label{sec:vision_sys}

\subsection{Scene Elements}
\label{sec:vision_sys_sceneelements}
The operator's visual feedback has four scene elements:

\subsubsection{Robot models}
Live URDF Spot models are rendered with joint states streamed from each robot via ROS~2. These models provide the operator with precise, always-visible kinematic state of each robot, including arm and gripper configurations.

\subsubsection{3D point clouds}
Frames from each robot's front-facing RGB-D cameras are combined into a point cloud in real time (Sec.~\ref{sec:visual_pipeline}). Both robots' point clouds are rendered as splats in the virtual view given the robot pose estimates (Sec.~\ref{sec:localization}).

\subsubsection{2D gripper-camera feeds}
Each gripper-camera feed is displayed as a floating panel near the operator's controller models. This close-up view compensates for cases where objects are occluded from the front-facing cameras, providing visual feedback for fine-grained grasping.

\subsubsection{Virtual gripper targets}
A virtual gripper is rendered at the commanded end-effector pose, distinct from the robot model's gripper. As network latency and robot dynamics delay the robot's physical response to the operator's input, this target gives immediate visual feedback of the operator's intended gripper pose; the robot model's gripper approaches the virtual gripper as the robot matches the target pose.

Of the four elements above, the robot models and virtual gripper targets are rendered directly from state data with negligible cost. The 2D gripper-camera feeds and 3D point clouds both originate from the vision processing pipeline.

\subsection{Vision Processing Pipeline}
\label{sec:visual_pipeline}

This pipeline must (1) have low latency, (2) register both robots' point clouds in a shared frame, and (3) complete sparse, noisy depth measurements into dense metric point clouds.

\begin{figure}[b!]
    \centering
    %\vspace{-1em}
    \includegraphics[width=0.85\linewidth]{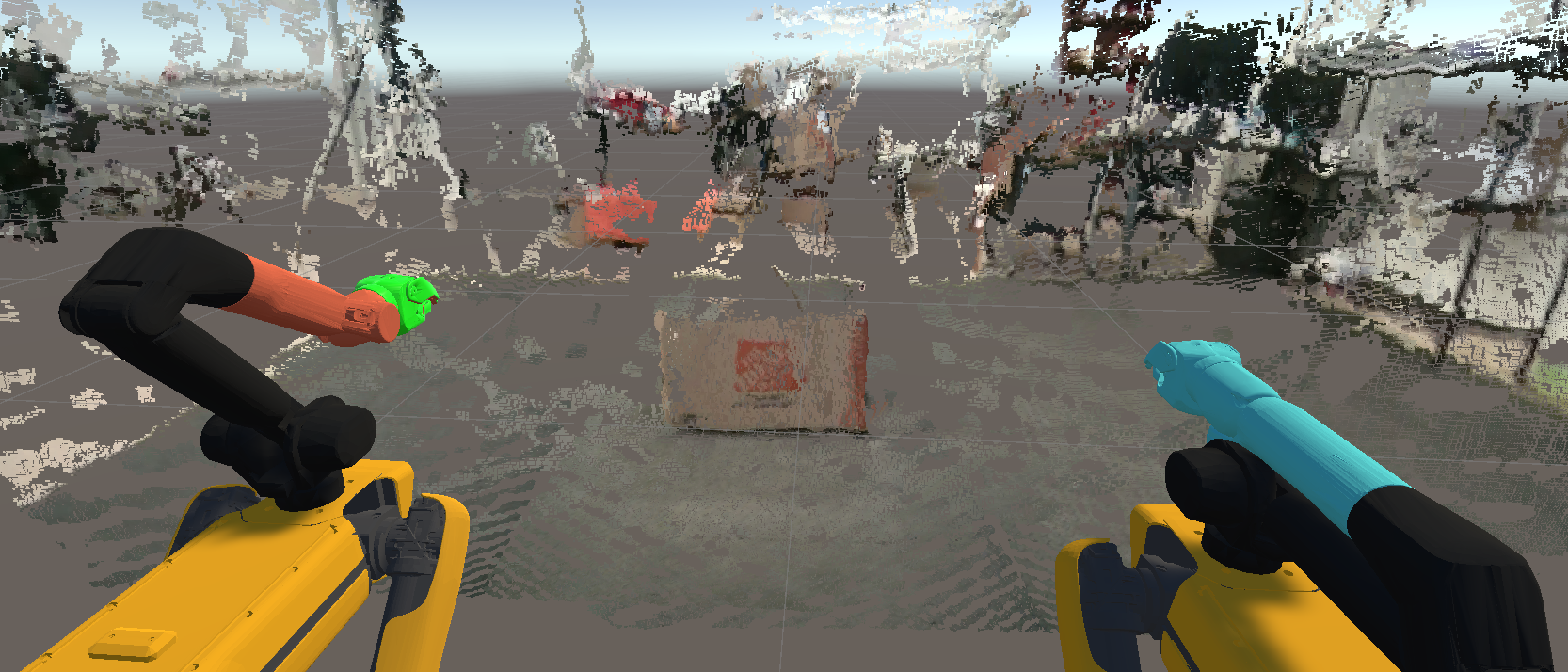}\\
    \includegraphics[width=0.85\linewidth]{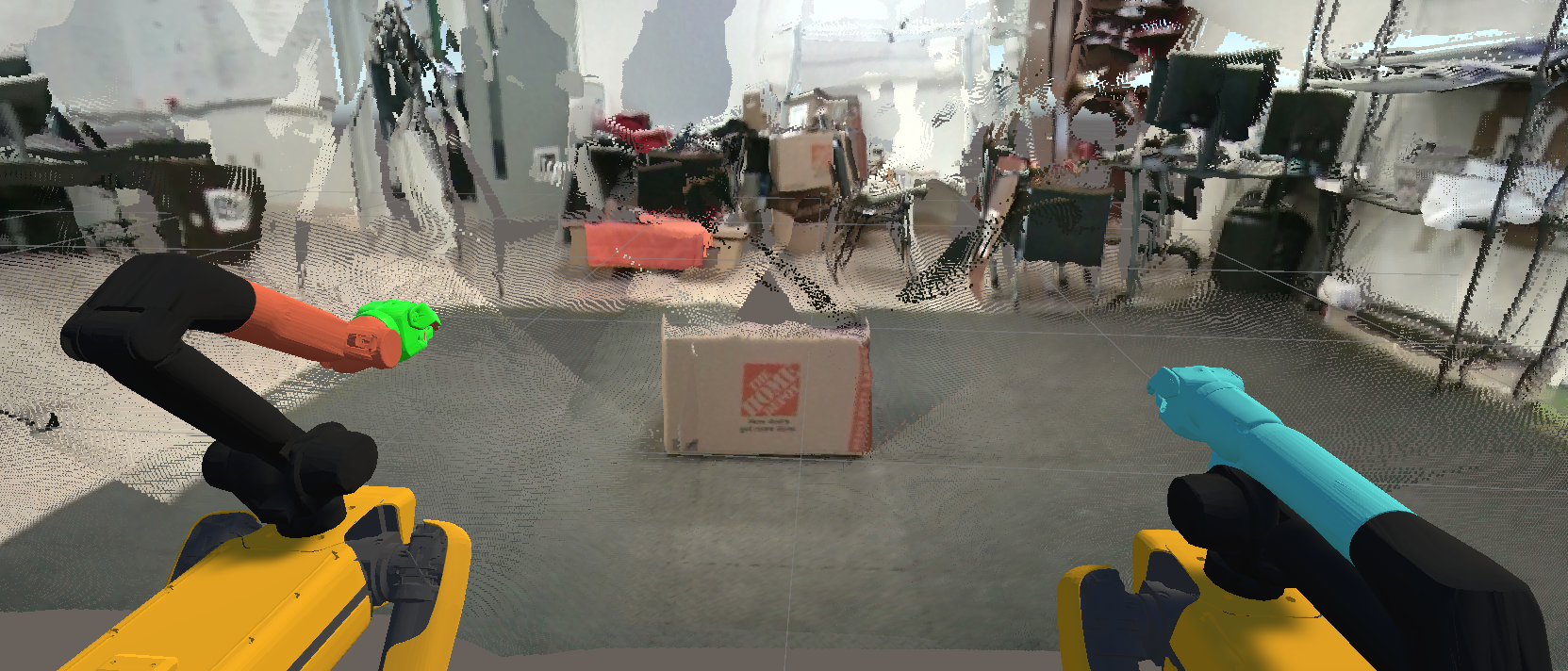}
    \caption{
        \textbf{Shared 3D scene before (above) and after (below) real-time depth completion.} Completion improves scene coverage and makes the point clouds from both robots' cameras more coherent.
    }
    \label{fig:depth_completion_comparison}
\end{figure}

% begin evaluation task figure
\newcommand{\taskimg}[3]{%
  \begin{tikzpicture}
    \node[anchor=south west,inner sep=0] (img) at (0,0)
      {\includegraphics[width=#1]{#2}};
    \begin{scope}[x={(img.south east)},y={(img.north west)}]
      \node[anchor=south west, fill=black, fill opacity=0.55, text opacity=1,
            text=white, font=\sffamily\fontsize{6}{7}\selectfont,
            inner sep=2pt, rounded corners=1.5pt]
        at (0.02,0.02) {#3};
    \end{scope}
  \end{tikzpicture}%
}
\begin{figure*}[t]
    \centering
    \newlength{\taskimggap}\setlength{\taskimggap}{3pt}%
    \newlength{\taskimgwidth}\setlength{\taskimgwidth}{\dimexpr(\textwidth - 4\taskimggap)/5\relax}%
    \taskimg{\taskimgwidth}{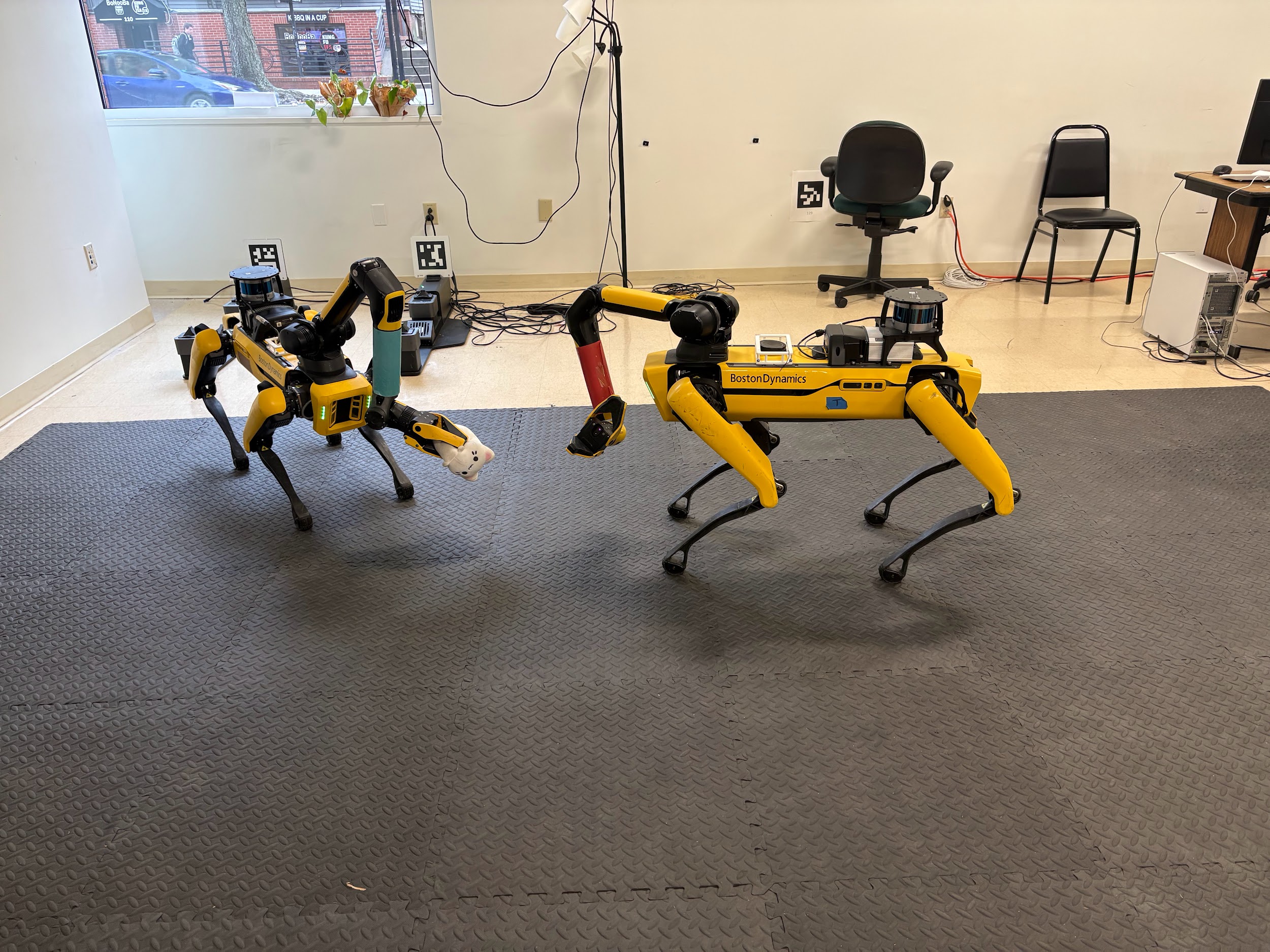}{1. Plush Toy Handoff}%
    \hspace{\taskimggap}%
    \taskimg{\taskimgwidth}{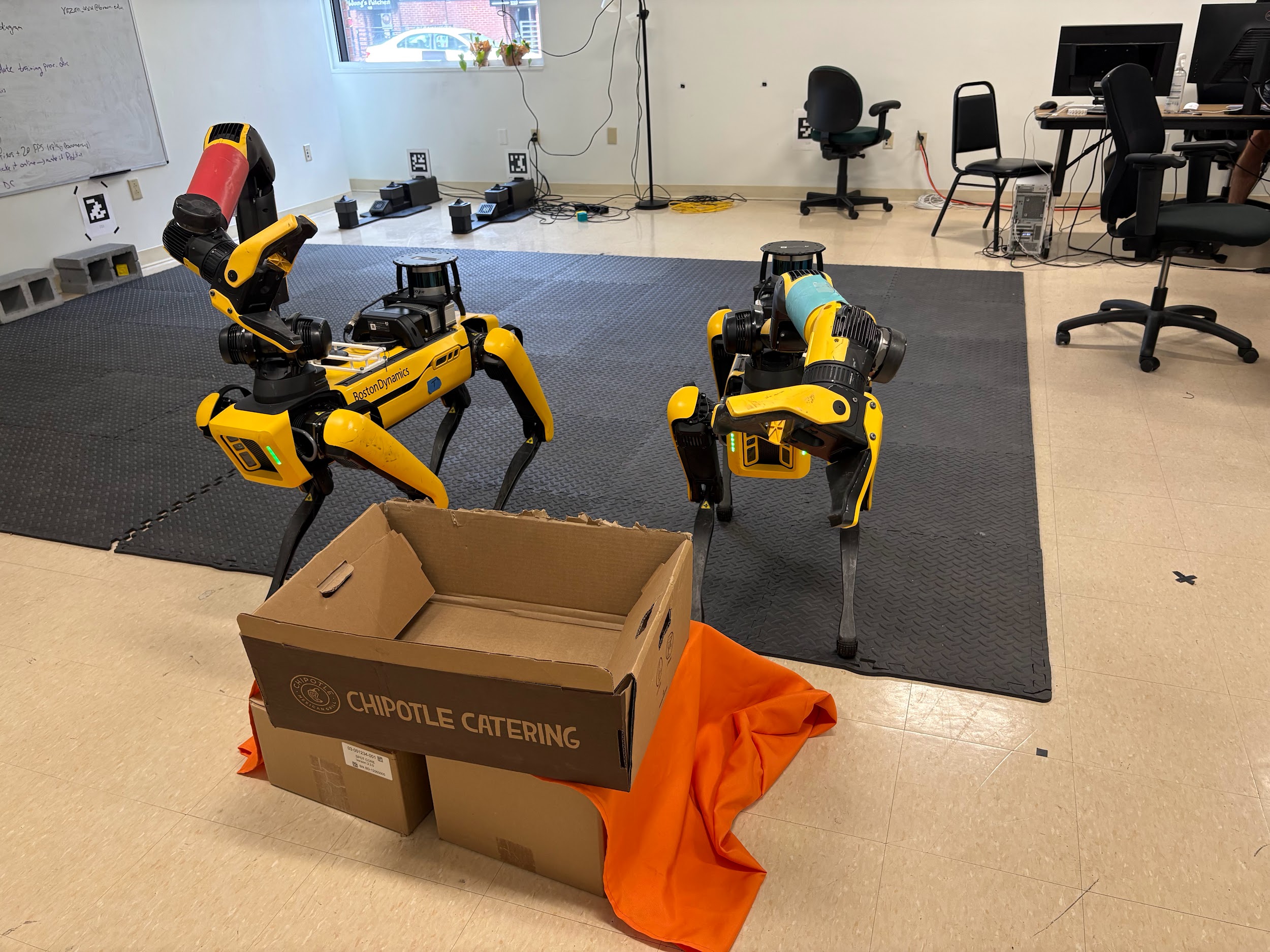}{2. Box Pick \& Place}%
    \hspace{\taskimggap}%
    \taskimg{\taskimgwidth}{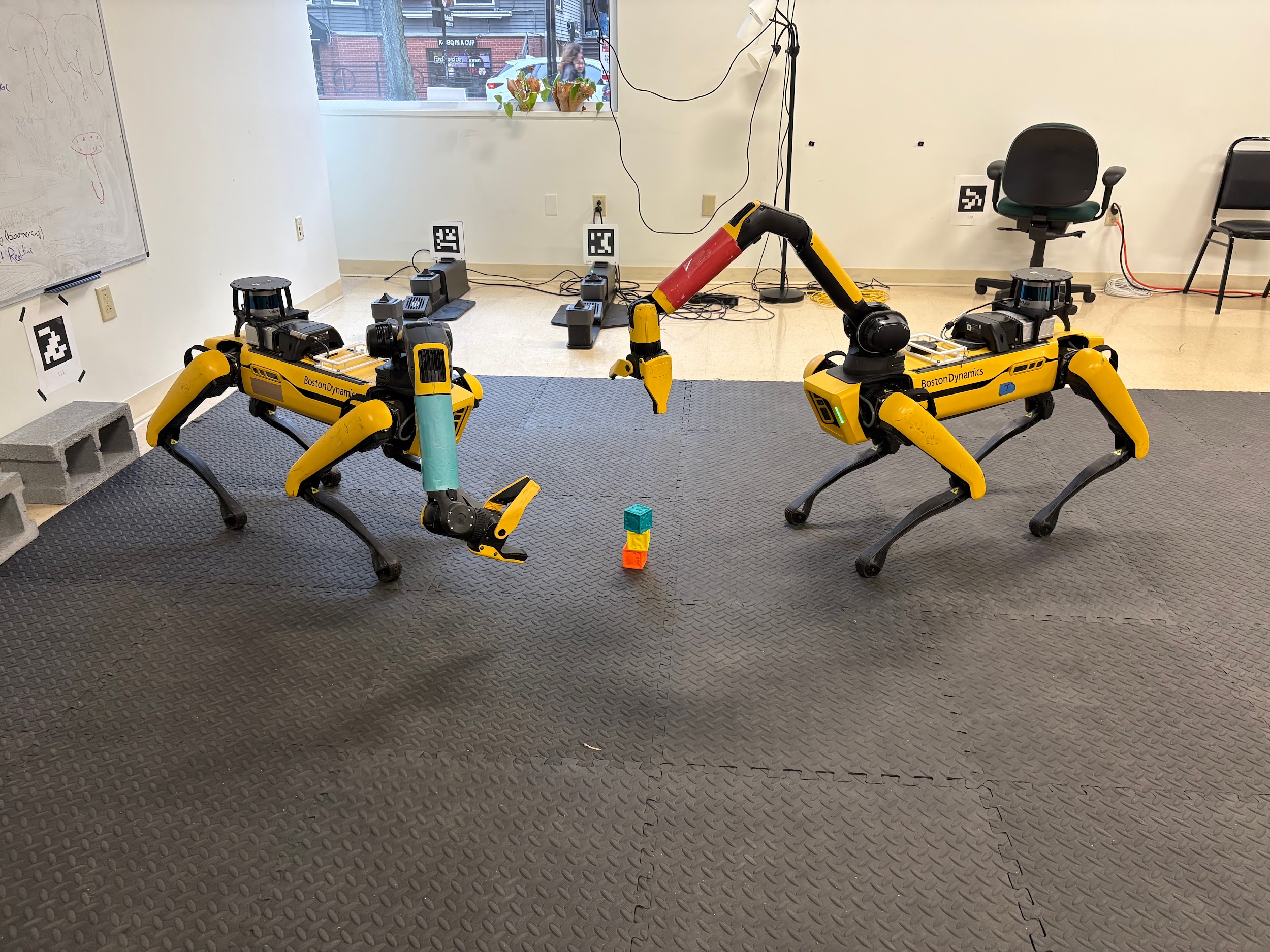}{3. Cube Stacking}%
    \hspace{\taskimggap}%
    \taskimg{\taskimgwidth}{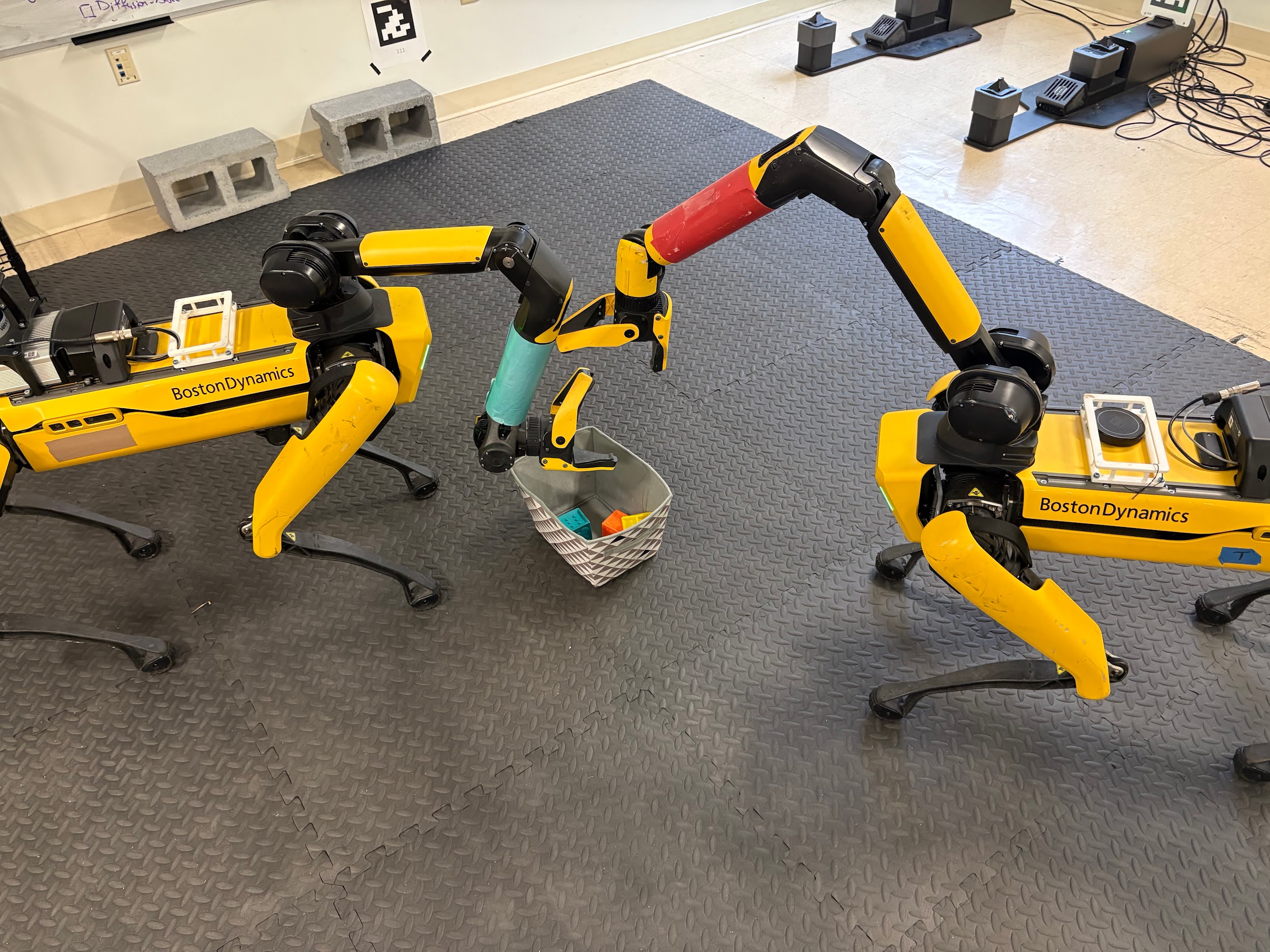}{4. Cube Collection}%
    \hspace{\taskimggap}%
    \taskimg{\taskimgwidth}{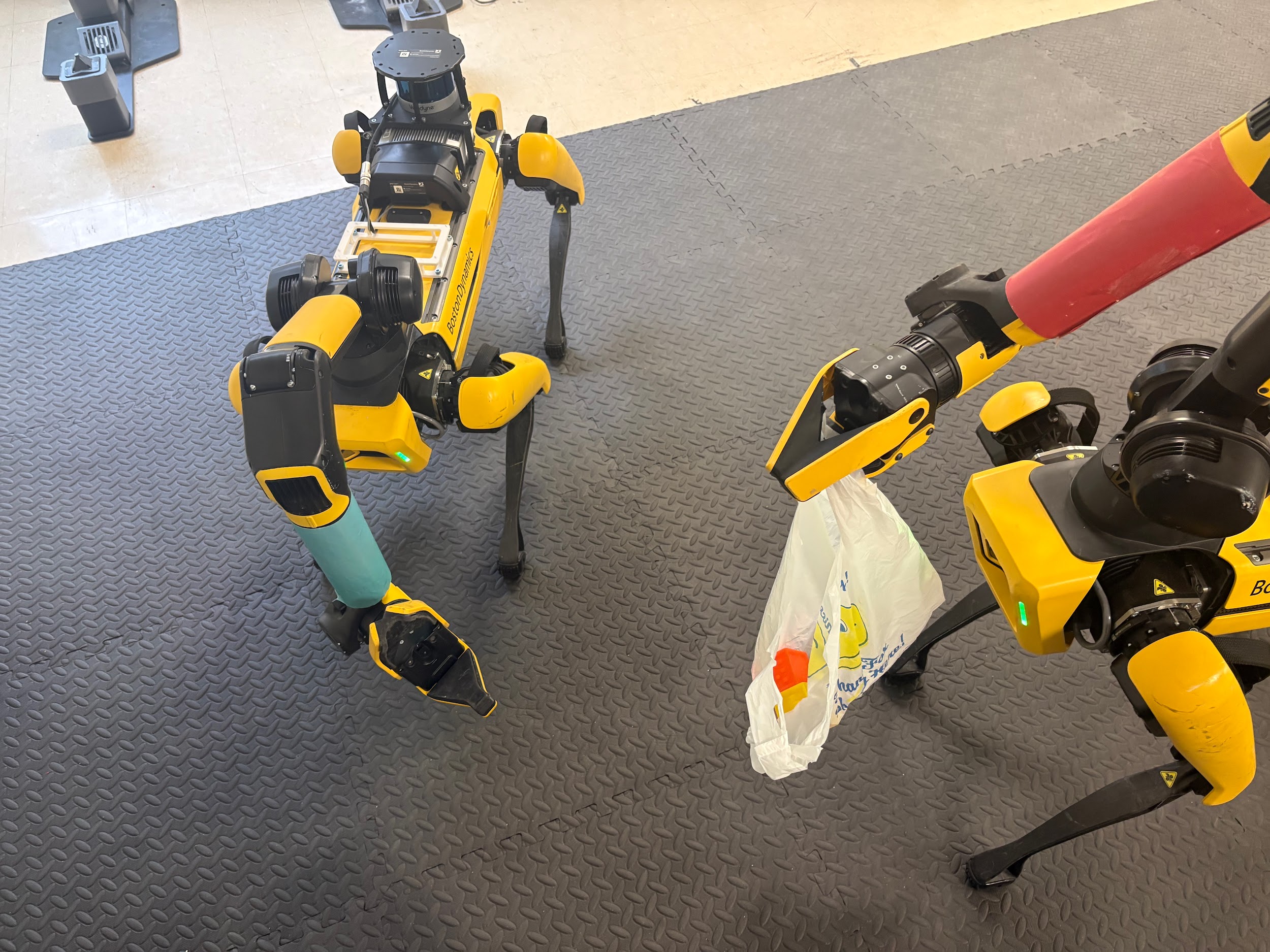}{5. Bag Filling}%

    \vspace{3pt}

    \taskimg{\taskimgwidth}{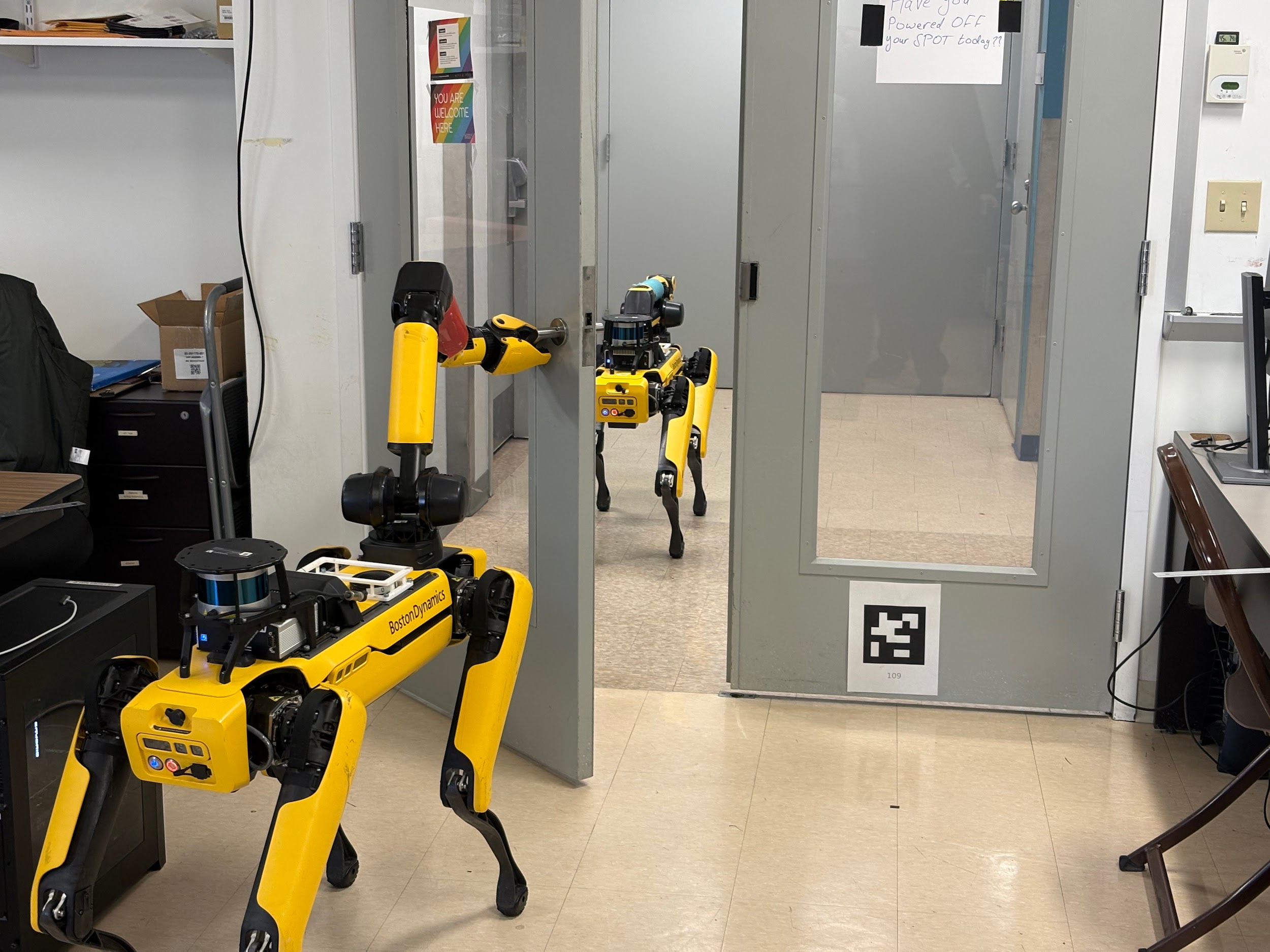}{6. Door Passthrough}%
    \hspace{\taskimggap}%
    \taskimg{\taskimgwidth}{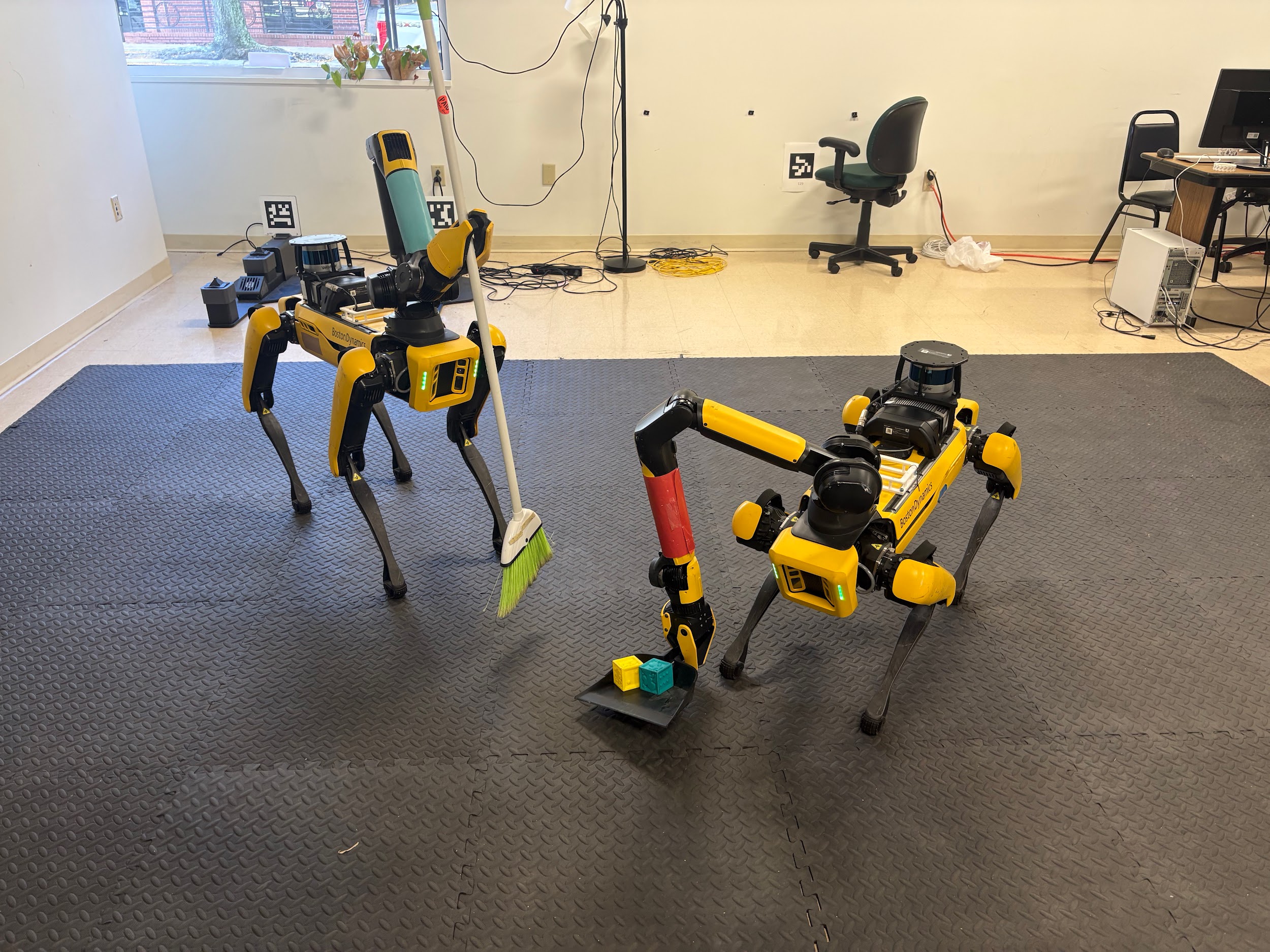}{7. Dust-pan Sweep}%
    \hspace{\taskimggap}%
    \taskimg{\taskimgwidth}{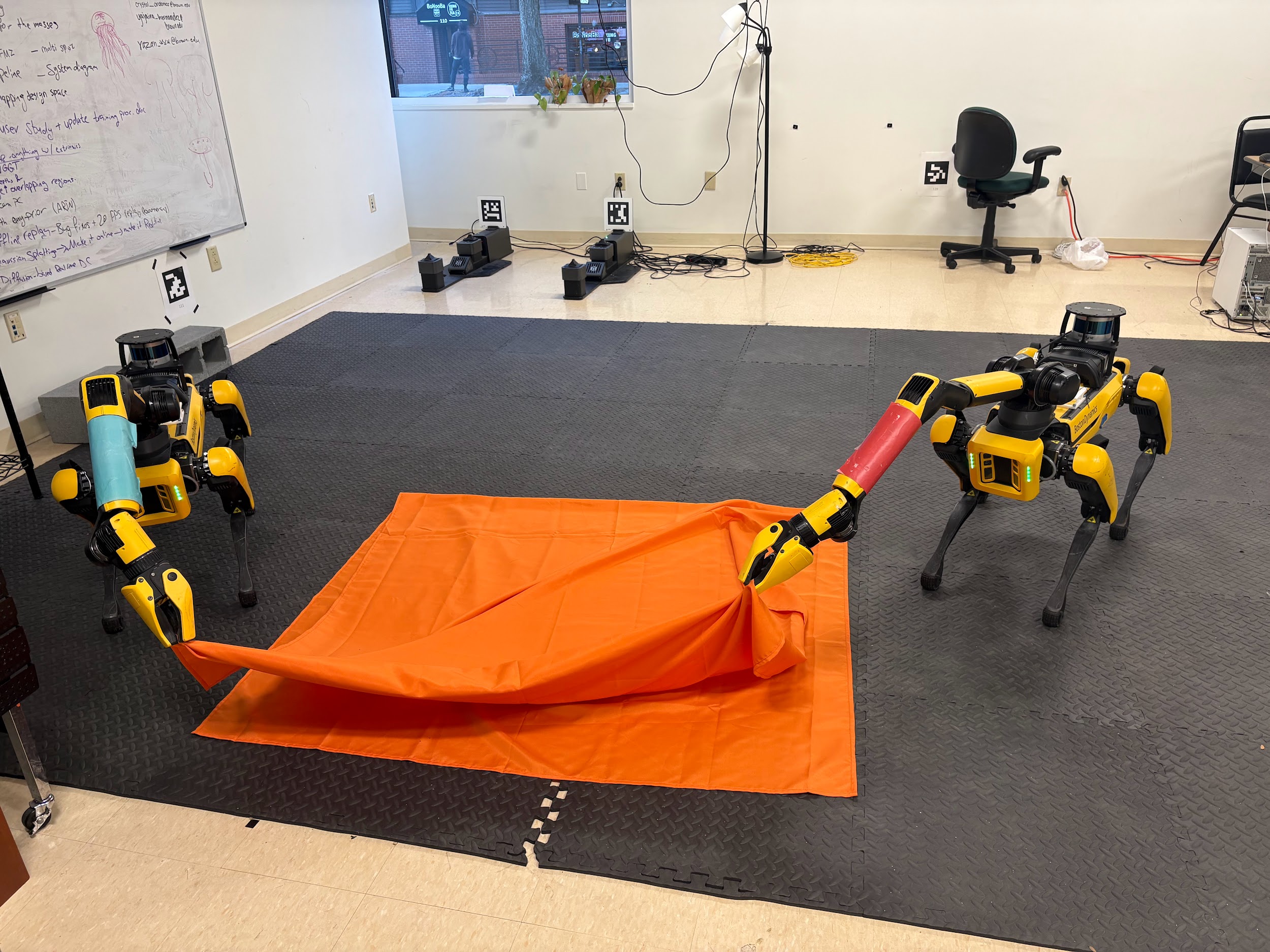}{8. Bedsheet Folding}%
    \hspace{\taskimggap}%
    \taskimg{\taskimgwidth}{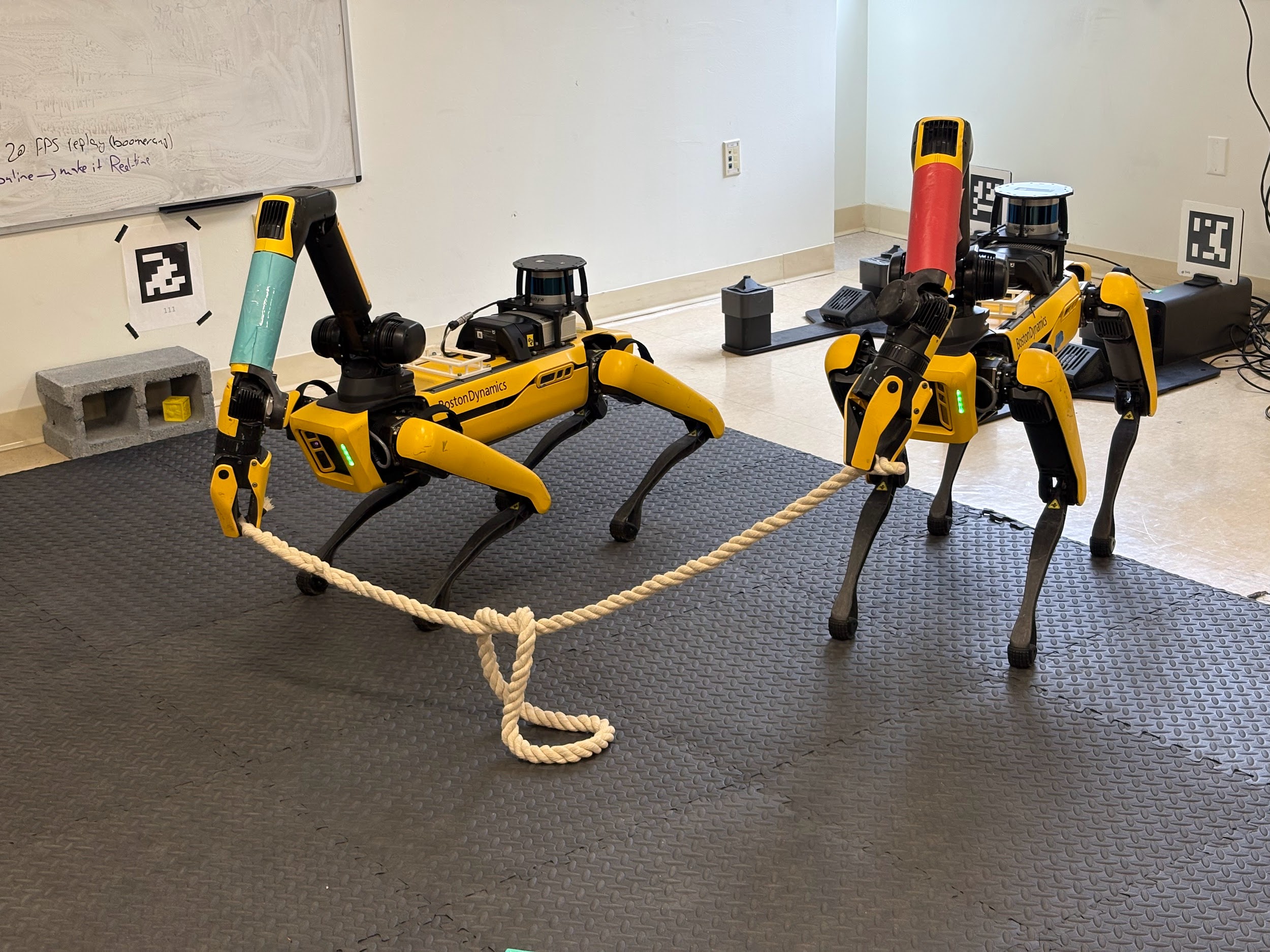}{9. Rope Tying}%
    \caption{\textbf{Nine evaluation tasks for dual-robot teleoperation.} Tasks require bimanual manipulation (1, 2, 5, 9), address perceptual blind spots (3, 4), or demand spatial context awareness (6, 7, 8). All require coordinated control of two mobile manipulators.}
    \label{fig:evaluation_tasks}
    %\vspace{-1em}
\end{figure*}
% end evaluation task figure

\subsubsection{Latency reduction} 

High-bandwidth (i.e., large payload) RGB-D streams are processed by SpotObserver in threads separate from Unity's render loop: for each robot, dedicated threads fetch RGB-D data and run depth completion, while the main rendering thread transfers completed data to Unity. SpotObserver shares GPU resources with Unity through CUDA-DX12 interoperability, avoiding CPU-GPU copies when moving data into the rendering context.

For our experiments, the robots communicated with the VR client and ROS~2 server over a dedicated Wi-Fi network. The robots streamed their RGB-D data at $640{\times}480$ resolution. The system achieved \SI{71}{\fps} headset rendering with a \SI{4.3}{\fps} scene update rate, limited primarily by Wi-Fi bandwidth. Latency is dominated by sensor-frame delivery over the network rather than by depth completion or rendering (Tab.~\ref{tab:sob_latency}).

\subsubsection{Relative 3D transform} 
\label{sec:localization}
Each robot's built-in localization estimates its base pose in a world frame, giving an initial inter-robot transform. Because these estimates are insufficient for point-cloud alignment, we refine the transform with an iterative closest point (ICP) correction computed from both robots' LiDAR data. Robot localization updates continuously, while the ICP correction is recomputed every 4 seconds.

\subsubsection{Metric depth completion} SpotObserver uses PromptDA for depth completion to clean up noisy and sparse depth-sensor data (Fig.~\ref{fig:depth_completion_comparison}). To improve completion quality, we apply nearest-neighbor interpolation to prefill sparse depth measurements before feeding them to the model. Both prefilling and model inference run on the GPU via CUDA and ONNX Runtime.

\begin{table}[t]
    \centering
    \caption{\textbf{Average SpotObserver latency per stage.}}
    \begin{tabular}{c c c}
    \toprule
    Sensor-frame delivery & Depth completion & Render \\
    \midrule
    \SI{230}{\milli\second} & \SI{131}{\milli\second} & \SI{14}{\milli\second} \\
    \bottomrule
    \end{tabular}
    \label{tab:sob_latency}
    %\vspace{-2em}
\end{table}

\section{Control}
\label{sec:control_design}

\subsection{Design}
Controlling two mobile manipulators simultaneously is a challenge in human attention allocation and interface design. The operator must manage each robot's base and arm controls, and the exocentric camera, resulting in five simultaneous control channels.
The Meta Quest 3 provides two hand controllers, each equipped with four buttons, a thumbstick, and 6-DoF tracking.
To effectively map the available controls to the required control channels, we designed a mode-switching architecture that presents operators with task-relevant control combinations.

GHOST separates \textit{robot selection} from \textit{control mode}. In single-robot operation, each hand controller can be assigned to one robot and placed in one of four task-level modes: \textit{Fly}, which repositions the exocentric camera; \textit{Drive}, which controls the robot base; \textit{Arm}, which maps controller pose to end-effector pose while providing gripper control; and \textit{Arm-Drive}, which combines base and arm control on one controller. In addition, GHOST provides a distinct \textit{Dual-Robot} mode for coordinated operation. In this mode, base commands are applied to a shared formation pivot so both robots move synchronously and maintain their relative configuration, while arm commands remain independently controlled by the operator's two hands. 
This separation lets the operator switch between individual single-robot control and synchronized dual-robot motion using the limited inputs of two hand controllers.

\subsection{Mappings}

\subsubsection{Single-robot control}
For \textit{arm control}, GHOST maps the hand controller's relative 6-DoF motion to the robot end-effector. When the operator holds the controller trigger, the system registers the current controller and end-effector poses; subsequent controller motion is applied as a relative displacement from that initial pose. This latch mechanism lets operators release and resume control, covering the robot workspace without excessive physical motion. The virtual gripper target immediately shows the commanded pose in VR, while the physical arm follows at approximately \SI{10}{\hertz} through the Boston Dynamics API and IK solver.

For \textit{base control}, thumbstick displacement is sent as translational and rotational velocity commands to the selected robot. For \textit{camera control}, the operator's head pose is tracked relative to a movable virtual base frame. Thumbstick input in \textit{Fly} mode translates or rotates this base frame, allowing the operator to navigate the exocentric scene without needing to physically walk through the workspace.

\subsubsection{Dual-robot synchronous control}
\label{sec:sync_formation_control}
To support coordinated navigation, such as jointly carrying a shared object, we implement a synchronous navigation mode. 
Upon activation, the initial base poses are recorded as $T_{w \to i}(0) \in SE(2)$ for $i=1,\ldots,n$ robots. The formation pivot position is the centroid of all robot positions
\(
\mathbf{p}_{\text{pivot}}(0) = \frac{1}{n} \sum_{i=1}^{n} \mathbf{p}_i(0)
\)
with yaw defined as the circular mean of robot yaw $\theta_i(0)$:
\begin{equation}
\bar{\theta}_{\text{yaw}}(0) = \text{atan2}\left(\frac{1}{n}\sum_{i=1}^{n} \sin(\theta_i(0)), \frac{1}{n}\sum_{i=1}^{n} \cos(\theta_i(0))\right).
\end{equation}

Each robot's offset from the pivot is recorded as $T_{\text{piv} \to i} = T_{\text{piv} \to w}(0) \cdot T_{w \to i}(0)$. When the operator issues a navigation command $\Delta T \in SE(2)$, the pivot updates to $T_{w \to \text{piv}}(t) = T_{w \to \text{piv}}(t-1) \cdot \Delta T$, and each robot receives its target waypoint $T_{w \to i}^{\text{target}}(t) = T_{w \to \text{piv}}(t) \cdot T_{\text{piv} \to i}$ via the Boston Dynamics API at \SI{10}{\hertz}. Empirically, this maintains sufficient formation accuracy for cooperative carrying over Wi-Fi. 
Arm controls remain independent across robots in this mode, since this independence aligns with humans' natural bimanual coordination.

\section{Evaluation}
\vspace{-0.1cm}

\subsection{Evaluation Design}

We evaluate GHOST on nine dual-robot tasks (Fig.~\ref{fig:evaluation_tasks}) spanning three categories:

\begin{itemize}
    \item \textbf{Bimanual manipulation} (Tasks 1, 2, 5, 9): Tasks requiring coordinated, simultaneous two-handed manipulation.
    \item \textbf{Perceptual blind spots} (Tasks 3, 4): Tasks that benefit from multi-viewpoint observations to reduce occlusion and improve scene coverage.
    \item \textbf{Spatial context awareness} (Tasks 6, 7, 8): Tasks that require reasoning about the relative spatial configuration of multiple robots and the environment.
\end{itemize}

Together, these tasks cover a range of coordination scenarios where multi-robot systems are required or provide practical advantages, while still capturing core single-robot capabilities.

\begin{figure}[b]
    \centering
    %\vspace{-1em}
    \includegraphics[width=0.7\linewidth]{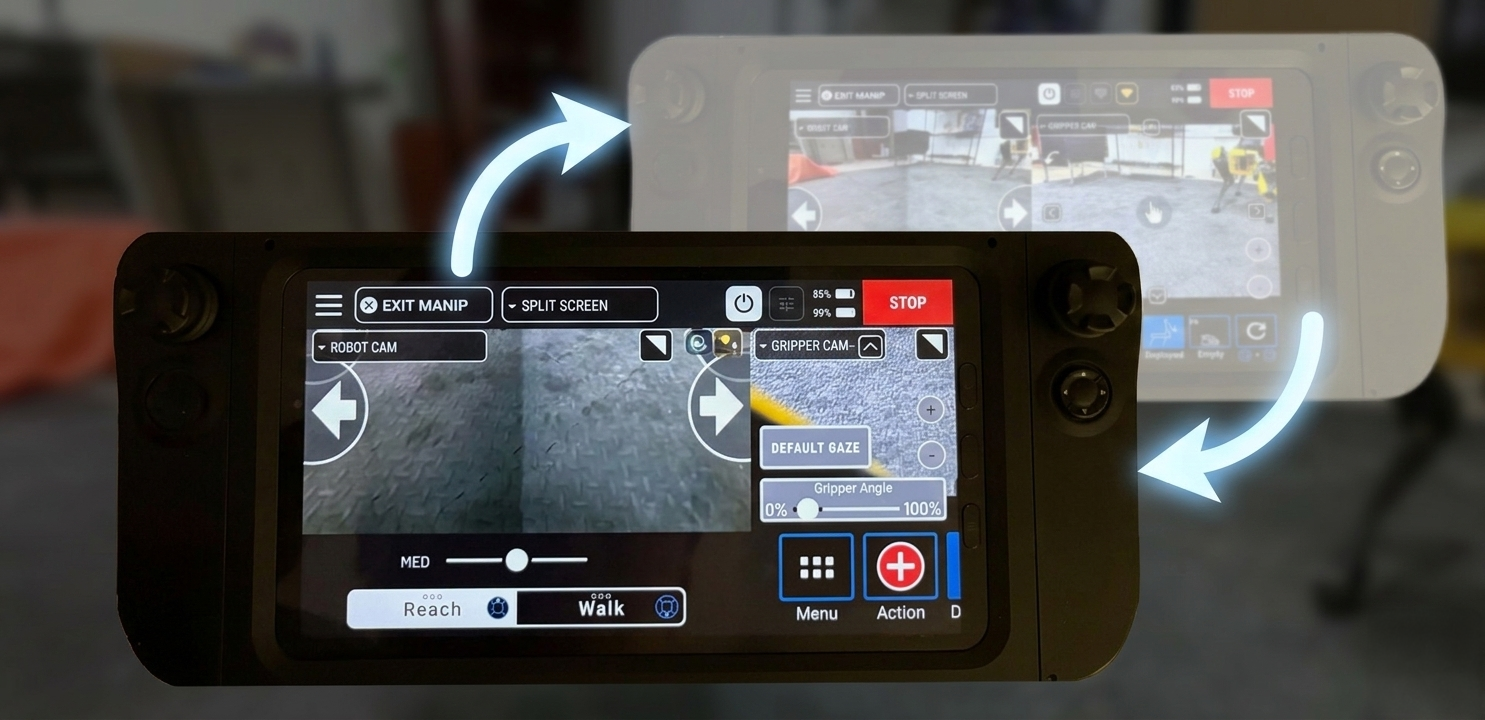}
    \caption{\textbf{Tablet interface.} Teleoperating two robots requires using two tablets, typically by switching between them.}
    \label{fig:tablet_interface}
\end{figure}

We compare to the official Boston Dynamics tablet interface: an egocentric, velocity-control baseline that requires two tablets (Fig.~\ref{fig:tablet_interface}) for simultaneous dual-robot teleoperation.

\begin{figure}[t]
    \centering
    \vspace{-1em}
    \includegraphics[width=0.885\linewidth]{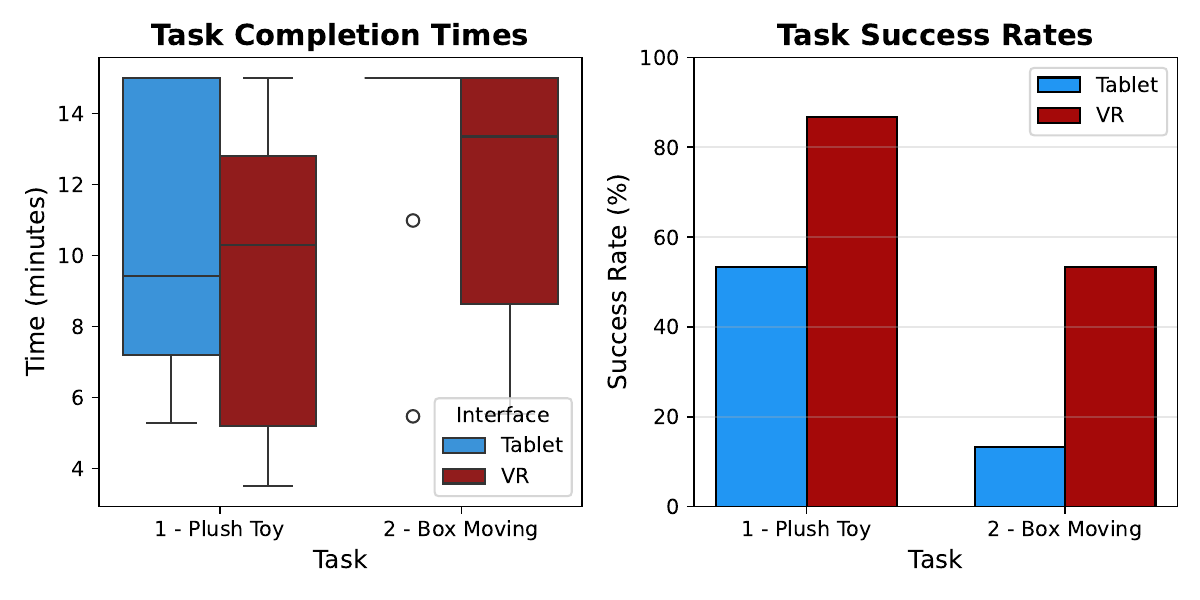}
    \vspace{-0.5em}
    \caption{
        \textbf{Novice task completion times and success rates.}
    }
    \label{fig:novice_success}
    %\vspace{-1.5em}
\end{figure}

\begin{figure}[b]
    \centering
    %\vspace{-1.5em}
    \includegraphics[width=0.6\linewidth]{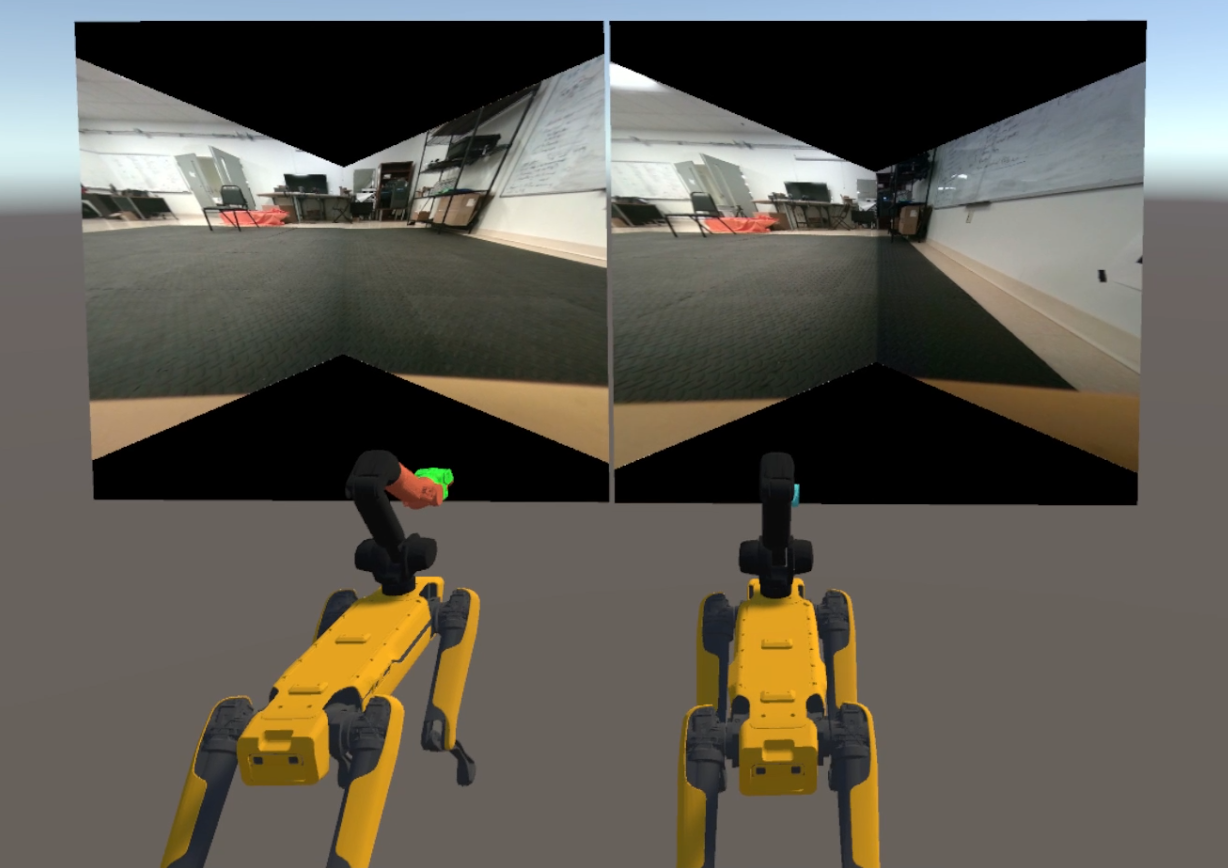}
    \caption{\textbf{Ablation study:} Operator view in RGB-only GHOST.}
    \label{fig:vr_ablation_rgb_interface}
\end{figure}

%% begin expert results
\begin{figure*}[t]
    \centering
    \vspace{-1em}
    \begin{minipage}[t]{0.7\linewidth}
        \vspace{0pt} % Makes 't' work
        \includegraphics[width=\linewidth]{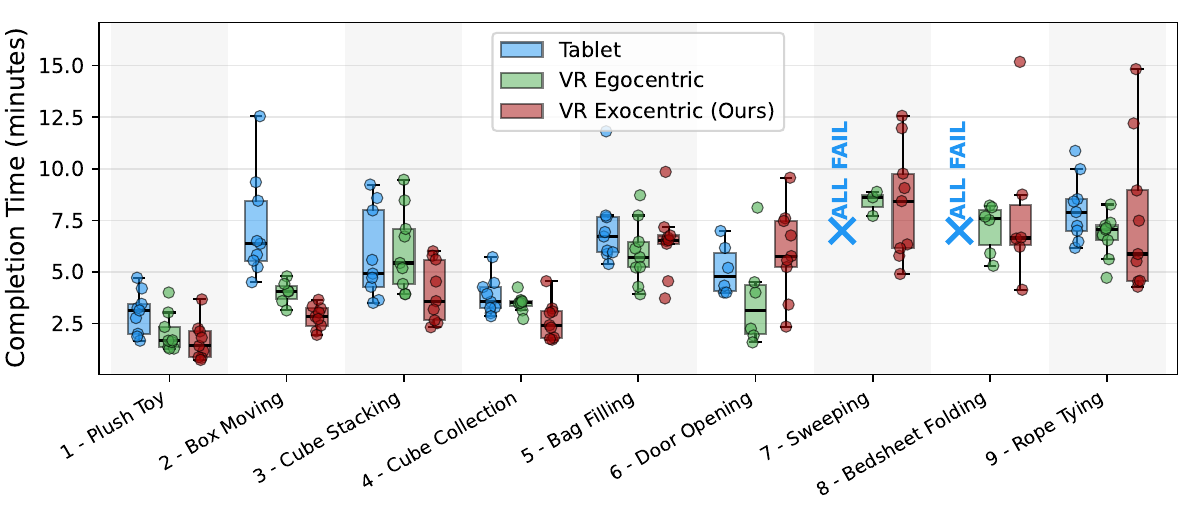}
    \end{minipage}%
    \hspace{.5em}
    \begin{minipage}[t]{0.28\linewidth}
        \vspace{0pt}
        \caption{\textbf{Expert task completion times across interfaces.} Full GHOST is faster and more reliable than the tablet baseline on most tasks, while the RGB-only GHOST ablation isolates the effect of direct 6-DoF control without the point-cloud-based exocentric scene. Two tasks were infeasible with the tablet. RGB-only GHOST generally falls between the tablet and full GHOST; when point-cloud quality is poor, it can match or exceed full GHOST.}
        \label{fig:expert_time}
    \end{minipage}
\end{figure*}
%%% end expert results

\subsection{Novice User Study}

To evaluate system usability with novice operators, we conducted a within-subjects study comparing the interfaces.

\textbf{Participants:} We recruited 15 novice participants (mean age $21.0 \pm 2.4$ years). Participants had minimal to no prior experience interacting with VR systems (mean rating $1.53 \pm 0.74$ on a 5-point scale from 1=never used to 5=use daily) or robotic systems ($1.87 \pm 0.83$).
Participants reported slightly more exposure to joystick-based games ($2.17 \pm 1.03$).

\textbf{Procedure:} 
Novice participants attempted two of the nine evaluation tasks: plush toy handoff and box pick-and-place. We selected these tasks because they are simple enough for first-time users while still requiring dual-robot coordination. Before each interface condition, participants received 10--15 minutes of familiarization and practiced picking up a cube from the ground. They then attempted both tasks with both GHOST and the tablet, while we recorded task success and completion time. Interface order was counterbalanced: 8 participants used GHOST first, and 7 used the tablet first.

\textbf{Metrics:} We measured task success rate and completion time. To assess perceived usability, participants also completed the System Usability Scale (SUS) after using each interface.

\textbf{Results:} Our system achieved 1.6$\times$ the tablet's success rate on plush toy handoff and 4$\times$ on box pick-and-place, with comparable or faster completion times among successful trials (Fig.~\ref{fig:novice_success}). On the SUS, our interface scored higher than the tablet on average ($64.8 \pm 19.2$ vs.\ $52.8 \pm 22.4$ out of 100). This difference was not statistically significant (Wilcoxon signed-rank test, $p = 0.16$, $d_z = 0.34$); we note the low $n$ in this preliminary study.
In feedback, participants cited point-cloud reconstruction quality and motion-induced discomfort as the main drawbacks of GHOST, while praising its intuitive control.

\subsection{Expert User Study}

We conducted an expert study to evaluate the capability of GHOST on the full set of dual-robot teleoperation tasks.

\textbf{Interfaces:} We tested three interfaces: full GHOST, GHOST with RGB-only camera feeds (Fig.~\ref{fig:vr_ablation_rgb_interface}), and the official tablet interface (Fig.~\ref{fig:tablet_interface}). RGB-only GHOST keeps the VR interface but removes the point-cloud 3D scene, separating the effect of 6-DoF control from 3D reconstruction.

\textbf{Participants and Procedure:}
Three authors experienced with all interfaces served as expert participants. Each participant attempted all nine tasks with each interface, performing three primary trials with up to two retries after failures.

\textbf{Metrics:} We measured task success, which defines feasibility, and completion time. If an expert failed more than three of five attempts, the task was deemed infeasible for that expert and excluded from timing comparisons.

\begin{table}[t]
    \centering
    \caption{\textbf{Expert task feasibility.}}
    \begin{tabular}{l ccc}
    \toprule
         &  GHOST (Ours) & GHOST (RGB-only) & Tablet \\
         \midrule
        1 - Plush Toy Handoff & \greentick & \greentick & \greentick \\
        2 - Box Pick \& Place & \greentick & 2/3        & \greentick \\
        3 - Cube Stacking    & \greentick & \greentick & \greentick \\
        4 - Cube Collection  & \greentick & \greentick & \greentick \\
        5 - Bag Filling      & \greentick & \greentick & \greentick \\
        6 - Door Passthrough  & \greentick & 2/3        & 2/3 \\
        7 - Dust-pan Sweep    & \greentick & 1/3        & \redx \\
        8 - Bedsheet Folding & 2/3        & 2/3        & \redx \\
        9 - Rope Tying       & \greentick & \greentick & \greentick \\
    \bottomrule
    \end{tabular}
    \label{tab:expert_success}
    %\vspace{-1em}
\end{table}

\textbf{Results:} Full GHOST achieved the highest task completion reliability, with 26/27 feasible participant-task conditions (Tab.~\ref{tab:expert_success}): 8 of 9 tasks were feasible for all three experts, and bedsheet folding for 2 of 3. The tablet baseline achieved 20/27 conditions; dust-pan sweeping and bedsheet folding were infeasible for all experts.

On the seven tasks where the tablet succeeded, GHOST achieved a 1.47$\times$ average per-task speedup over the tablet (Fig.~\ref{fig:expert_time}). The largest gains were on box pick-and-place (2:48 vs.\ 7:09), plush toy handoff (1:39 vs.\ 3:00), cube stacking (4:01 vs.\ 5:50), and cube collection (2:39 vs.\ 3:49). GHOST was also slightly faster on bag filling and rope tying. Although GHOST was slower than the tablet on successful weighted-door passthrough trials (Task~6), this task was more often feasible with GHOST (3/3 vs.\ 2/3).

The RGB-only GHOST ablation achieved 22/27 feasible participant-task conditions. Its task completion reliability and completion times mostly lie between the tablet and full GHOST. RGB-only GHOST failed more on tasks requiring coordinated spatial reasoning, including box pick-and-place, weighted-door passthrough, dust-pan sweeping, and bedsheet folding. This suggests that direct 6-DoF control provides much of the timing benefit, while the reconstructed 3D scene improves reliability and spatial awareness when the point cloud is accurate.

\vspace{-0.15cm}
\subsection{Discussion}
While preliminary, the results show clear advantages of our approach over RGB-based, joystick-controlled dual-robot teleoperation.
We examine the key factors behind these differences: coordinated dual-arm control, direct pose control, predictive visual feedback, and 3D scene representation.

\subsubsection{Coordinated control}
VR maps each robot arm to the operator's corresponding hand, enabling simultaneous dual-arm control without mode switching.
This bimanual mapping was important for tightly coordinated tasks: synchronized grasping (Task~2), and bedsheet folding (Task~8).
In practice, tablet operators alternated between arms as simultaneous input across two tablets proved too demanding; this was a primary factor in the tablet's failure on Task~8.
However, simultaneous control helped little on sequential tasks like weighted-door passthrough (Task~6) or cube collection (Task~4).

\subsubsection{Intuitive high-precision and rotation control}
Direct hand-pose mapping provides an advantage over joystick-based velocity control, particularly for precise alignment (Task~3) and rotation-intensive tasks (Tasks~5, 7). Because VR maps hand pose directly to end-effector pose, fine spatial adjustments are immediate and natural, whereas velocity commands make precise placement and orientation cumbersome.

\subsubsection{Virtual gripper target reduces perceived latency}
The virtual gripper target provides a preview of the commanded pose, reducing perceived latency and improving confidence. However, during tool use (Task~7), it only previews gripper motion, not the tool tip, creating a perceptual mismatch that partially negates this benefit. Despite this, VR still outperforms the tablet for tool use, as velocity-based rotation makes compounded rotations at a tool tip far from the gripper difficult.

\subsubsection{3D representation tradeoffs}
A 3D point cloud representation enables precise spatial alignment (Task~3) and is a key driver of our system's performance: while the dual-robot setup partially mitigates the viewing-angle limitations of RGB-only systems, operators must still switch between camera views to achieve full alignment, whereas the reconstructed 3D scene provides a unified spatial reference for reasoning about object pose, inter-robot alignment, and spatial constraints. This advantage is more evident in task completion reliability. When comparing RGB-only GHOST versus full GHOST, the 3D scene yielded only marginal gains in completion time but raised completions from 22/27 to 26/27, suggesting it helps operators with coordinated spatial reasoning that RGB views alone make difficult. However, the benefit depends on reconstruction quality: accurate depth completion provided useful spatial cues, while inaccurate reconstruction resulted in reduced efficiency for tasks that involve deformable or cluttered objects (Tasks~5 and~9). This likely drives the high variability: in some cases RGB-only GHOST matched or exceeded full GHOST, indicating that an inaccurate 3D scene can be worse than none at all.

\section{Conclusions}
We present GHOST, an exocentric VR system that enables a single operator to perform low-level dual-robot mobile manipulation using only onboard sensing. Across novice and expert studies, GHOST improved success rate and typically also completion time relative to an egocentric tablet baseline, and enabled tasks that were infeasible with the baseline.

Our user study has limitations. First, the expert participants are the developers of the system. Although they are familiar with both interfaces, bias is inevitable. Second, there were few novice participants, and each attempted only two tasks to remain manageable for new users. Thus, this study should be understood as a preliminary assessment of the system's capability. Future larger-scale studies with independent expert operators and a broader pool of novice participants could systematically characterize trends across teleoperation modalities.

Other promising directions include training behavior cloning models from the collected data, scaling to larger multi-robot teams, integrating autonomous assistance for shared control, and improving the 3D reconstruction pipeline to be more precise and consistent for better scene understanding.

\section*{Acknowledgments}
We thank Calvin Bauer, Are Oelsner, and Janeth Meraz for their work on an earlier version of GHOST~\cite{bauer2024ghost}. We also thank Ziyan Liu, Shiyu Yan, Arin Idhant, Simon Juknelis, Kamya Raman, Arib Syed, and Automne Petitjean for working on the project, and thank Ivy He, Mingxi Jia, and Gary Lvov for their valuable suggestions on the paper.

\balance
{\footnotesize
\bibliography{IEEEexample}

@IEEEtranBSTCTL{BSTcontrol,
  CTLuse_forced_etal = "yes",
  CTLmax_names_forced_etal = "6",
  CTLnames_show_etal = "1"
}

@inproceedings{swamy2020scaled_autonomy,
  title={Scaled Autonomy: Enabling Human Operators to Control Robot Fleets},
  author={Swamy, Gokul and Reddy, Siddharth and Levine, Sergey and Dragan, Anca D.},
  booktitle={2020 IEEE International Conference on Robotics and Automation (ICRA)},
  year={2020},
  organization={IEEE}
}

@inproceedings{whitney2018ros_reality,
  title={{ROS} Reality: A virtual reality framework using consumer-grade hardware for {ROS}-enabled robots},
  author={Whitney, David and Rosen, Eric and Ullman, Daniel and Phillips, Elizabeth and Tellex, Stefanie},
  booktitle={2018 IEEE/RSJ International Conference on Intelligent Robots and Systems (IROS)},
  year={2018},
  organization={IEEE}
}

@article{zick2024teleoperation,
  title={Teleoperation system for multiple robots with intuitive hand recognition interface},
  author={Zick, Lucas Alexandre and Martinelli, Dieisson and Schneider de Oliveira, Andr{\'e} and Cremer Kalempa, Vivian},
  journal={Scientific Reports},
  year={2024},
  publisher={Nature Publishing Group UK London}
}

@INPROCEEDINGS{10801345,
  author={Wilder-Smith, Maximum and Patil, Vaishakh and Hutter, Marco},
  booktitle={2024 IEEE/RSJ International Conference on Intelligent Robots and Systems (IROS)}, 
  title={Radiance Fields for Robotic Teleoperation}, 
  year={2024},
  doi={10.1109/IROS58592.2024.10801345}}

@article{qiu2025humanoid,
  title={Humanoid Policy $\sim$ Human Policy},
  author={Qiu, Ri-Zhao and Yang, Shiqi and Cheng, Xuxin and Chawla, Chaitanya and Li, Jialong and He, Tairan and Yan, Ge and Yoon, David J and Hoque, Ryan and Paulsen, Lars and others},
  journal={arXiv preprint arXiv:2503.13441},
  year={2025}
}

@inproceedings{tung2021collab_teleop,
  title={Learning multi-arm manipulation through collaborative teleoperation},
  author={Tung, Albert and Wong, Josiah and Mandlekar, Ajay and Mart{\'\i}n-Mart{\'\i}n, Roberto and Zhu, Yuke and Fei-Fei, Li and Savarese, Silvio},
  booktitle={2021 IEEE International Conference on Robotics and Automation (ICRA)},
  year={2021},
  organization={IEEE}
}

@unpublished{bauer2024ghost,
  title        = {{GHOST} in the Robot: Virtual Reality Teleoperation for Mobile Manipulation},
  author       = {Calvin Bauer and Janeth Meraz and Are Oelsner and James Tompkin and Stefanie Tellex},
  year         = {2024},
  note         = {{MSc} Project},
}

@article{cheng2019learning,
  title={Learning depth with convolutional spatial propagation network},
  author={Cheng, Xinjing and Wang, Peng and Yang, Ruigang},
  journal={IEEE Transactions on Pattern Analysis and Machine Intelligence},
  year={2019},
  publisher={IEEE}
}

@inproceedings{ma2018sparse,
  title={Sparse-to-dense: Depth prediction from sparse depth samples and a single image},
  author={Ma, Fangchang and Karaman, Sertac},
  booktitle={2018 IEEE international conference on robotics and automation (ICRA)},
  year={2018},
  organization={IEEE}
}

@article{oquab2023dinov2,
  title={{DINOv2}: Learning Robust Visual Features without Supervision},
  author={Oquab, Maxime and Darcet, Timoth{\'e}e and Moutakanni, Th{\'e}o and Vo, Huy V. and Szafraniec, Marc and others},
  journal={Transactions on Machine Learning Research},
  year={2024}
}

@article{nconv,
  title={Confidence propagation through {CNNs} for guided sparse depth regression},
  author={Eldesokey, Abdelrahman and Felsberg, Michael and Khan, Fahad Shahbaz},
  journal={IEEE Transactions on Pattern Analysis and Machine Intelligence},
  year={2019},
  publisher={IEEE}
}

@inproceedings{bpnet,
  title={Bilateral Propagation Network for Depth Completion},
  author={Tang, Jie and Tian, Fei-Peng and An, Boshi and Li, Jian and Tan, Ping},
  booktitle={Proceedings of the IEEE/CVF Conference on Computer Vision and Pattern Recognition},
  year={2024}
}

@inproceedings{yang2024depth,
  title={{Depth Anything V2}},
  author={Yang, Lihe and Kang, Bingyi and Huang, Zilong and Zhao, Zhen and Xu, Xiaogang and Feng, Jiashi and Zhao, Hengshuang},
  booktitle={Advances in Neural Information Processing Systems},
  year={2024}
}

@inproceedings{viola2024marigolddc,
  title={{Marigold-DC}: Zero-Shot Monocular Depth Completion with Guided Diffusion},
  author={Viola, Massimiliano and Qu, Kevin and Metzger, Nando and Ke, Bingxin and Becker, Alexander and Schindler, Konrad and Obukhov, Anton},
  booktitle={Proceedings of the IEEE/CVF International Conference on Computer Vision},
  year={2025}
}

@inproceedings{zhang2023completionformer,
  title={{CompletionFormer}: Depth completion with convolutions and vision transformers},
  author={Zhang, Youmin and Guo, Xianda and Poggi, Matteo and Zhu, Zheng and Huang, Guan and Mattoccia, Stefano},
  booktitle={Proceedings of the IEEE/CVF Conference on Computer Vision and Pattern Recognition},
  year={2023}
}

@inproceedings{lin2025promptda,
  title={Prompting {Depth Anything} for {4K} resolution accurate metric depth estimation},
  author={Lin, Haotong and Peng, Sida and Chen, Jingxiao and Peng, Songyou and Sun, Jiaming and Liu, Minghuan and Bao, Hujun and Feng, Jiashi and Zhou, Xiaowei and Kang, Bingyi},
  booktitle={Proceedings of the IEEE/CVF Conference on Computer Vision and Pattern Recognition},
  year={2025}
}

@inproceedings{xiong2025vision_in_action,
  title={Vision in Action: Learning Active Perception from Human Demonstrations},
  author={Xiong, Haoyu and Xu, Xiaomeng and Wu, Jimmy and Hou, Yifan and Bohg, Jeannette and Song, Shuran},
  booktitle={Conference on Robot Learning (CoRL)},
  year={2025},
  organization={PMLR}
}

@inproceedings{iyer2025openteach,
  title={{OPEN TEACH}: A Versatile Teleoperation System for Robotic Manipulation},
  author={Iyer, Aadhithya and Peng, Zhuoran and Dai, Yinlong and Guzey, Irmak and Haldar, Siddhant and Chintala, Soumith and Pinto, Lerrel},
  booktitle={Conference on Robot Learning},
  year={2025},
  organization={PMLR}
}

@inproceedings{lin2024hato,
  title={Learning Visuotactile Skills with Two Multifingered Hands},
  author={Lin, Toru and Zhang, Yu and Li, Qiyang and Qi, Haozhi and Yi, Brent and Levine, Sergey and Malik, Jitendra},
  booktitle={2025 IEEE International Conference on Robotics and Automation (ICRA)},
  year={2025},
  organization={IEEE}
}

@inproceedings{zhao2023aloha,
  title={Learning Fine-Grained Bimanual Manipulation with Low-Cost Hardware},
  author={Zhao, Tony and Kumar, Vikash and Levine, Sergey and Finn, Chelsea},
  booktitle={Robotics: Science and Systems (RSS)},
  year={2023}
}

@inproceedings{cheng2025opentv,
  title={{Open-TeleVision}: Teleoperation with Immersive Active Visual Feedback},
  author={Cheng, Xuxin and Li, Jialong and Yang, Shiqi and Yang, Ge and Wang, Xiaolong},
  booktitle={Conference on Robot Learning},
  year={2025},
  organization={PMLR}
}

@inproceedings{wu2024gello,
  title={{GELLO}: A general, low-cost, and intuitive teleoperation framework for robot manipulators},
  author={Wu, Philipp and Shentu, Yide and Yi, Zhongke and Lin, Xingyu and Abbeel, Pieter},
  booktitle={2024 IEEE/RSJ International Conference on Intelligent Robots and Systems (IROS)},
  year={2024},
  organization={IEEE}
}

@article{honerkamp2025moma-teleop,
  title={Whole-body teleoperation for mobile manipulation at zero added cost},
  author={Honerkamp, Daniel and Mahesheka, Harsh and von Hartz, Jan Ole and Welschehold, Tim and Valada, Abhinav},
  journal={IEEE Robotics and Automation Letters},
  year={2025},
  publisher={IEEE}
}

@inproceedings{chuang2025active,
  title={Active vision might be all you need: Exploring active vision in bimanual robotic manipulation},
  author={Chuang, Ian and Lee, Andrew and Gao, Dechen and Naddaf-Sh, M-Mahdi and Soltani, Iman},
  booktitle={2025 IEEE International Conference on Robotics and Automation (ICRA)},
  year={2025},
  organization={IEEE}
}

@inproceedings{wang2025vggt,
  title={{VGGT}: Visual geometry grounded transformer},
  author={Wang, Jianyuan and Chen, Minghao and Karaev, Nikita and Vedaldi, Andrea and Rupprecht, Christian and Novotny, David},
  booktitle={Proceedings of the IEEE/CVF Conference on Computer Vision and Pattern Recognition},
  year={2025}
}

@inproceedings{keetha2025mapanything,
  title={{MapAnything}: Universal Feed-Forward Metric {3D} Reconstruction},
  author={Keetha, Nikhil and M{\"u}ller, Norman and Sch{\"o}nberger, Johannes and Porzi, Lorenzo and Zhang, Yuchen and Fischer, Tobias and Knapitsch, Arno and Zauss, Duncan and Weber, Ethan and Antunes, Nelson and others},
  booktitle={International Conference on 3D Vision (3DV)},
  year={2026},
  organization={IEEE}
}

@inproceedings{vit,
  title     = {An Image is Worth 16x16 Words: Transformers for Image Recognition at Scale},
  author    = {Alexey Dosovitskiy and Lucas Beyer and Alexander Kolesnikov and Dirk Weissenborn and Xiaohua Zhai and Thomas Unterthiner and Mostafa Dehghani and Matthias Minderer and Georg Heigold and Sylvain Gelly and Jakob Uszkoreit and Neil Houlsby},
  booktitle = {International Conference on Learning Representations (ICLR)},
  year      = {2021}
}

@inproceedings{DBLP:journals/corr/abs-2112-10752,
  title={High-Resolution Image Synthesis with Latent Diffusion Models},
  author={Rombach, Robin and Blattmann, Andreas and Lorenz, Dominik and Esser, Patrick and Ommer, Bj{\"o}rn},
  booktitle={Proceedings of the IEEE/CVF Conference on Computer Vision and Pattern Recognition},
  year={2022}
}

@article{ozdamar2022shared,
  title={A shared autonomy reconfigurable control framework for telemanipulation of multi-arm systems},
  author={Ozdamar, Idil and Laghi, Marco and Grioli, Giorgio and Ajoudani, Arash and Catalano, Manuel G and Bicchi, Antonio},
  journal={IEEE Robotics and Automation Letters},
  year={2022},
  publisher={IEEE}
}

@article{dass2024telemoma,
  title={{TeleMoMa}: A modular and versatile teleoperation system for mobile manipulation},
  author={Dass, Shivin and Ai, Wensi and Jiang, Yuqian and Singh, Samik and Hu, Jiaheng and Zhang, Ruohan and Stone, Peter and Abbatematteo, Ben and Mart{\'\i}n-Mart{\'\i}n, Roberto},
  journal={arXiv preprint arXiv:2403.07869},
  year={2024}
}

@inproceedings{fu2024mobilealoha,
  title={Mobile {ALOHA}: Learning Bimanual Mobile Manipulation using Low-Cost Whole-Body Teleoperation},
  author={Fu, Zipeng and Zhao, Tony Z. and Finn, Chelsea},
  booktitle={Conference on Robot Learning},
  year={2025},
  organization={PMLR}
}

@inproceedings{mandlekar2018roboturk,
  title={{RoboTurk}: A crowdsourcing platform for robotic skill learning through imitation},
  author={Mandlekar, Ajay and Zhu, Yuke and Garg, Animesh and Booher, Jonathan and Spero, Max and Tung, Albert and Gao, Julian and Emmons, John and Gupta, Anchit and Orbay, Emre and others},
  booktitle={Conference on Robot Learning},
  year={2018},
  organization={PMLR}
}

@inproceedings{sivakumar2022robotic_telekinesis,
  title={Robotic Telekinesis: Learning a Robotic Hand Imitator by Watching Humans on {YouTube}},
  author={Sivakumar, Aravind and Shaw, Kenneth and Pathak, Deepak},
  booktitle={Robotics: Science and Systems (RSS)},
  year={2022}
}

@inproceedings{qin2023anyteleop,
  title={{AnyTeleop}: A General Vision-Based Dexterous Robot Arm-Hand Teleoperation System},
  author={Qin, Yuzhe and Yang, Wei and Huang, Binghao and Van Wyk, Karl and Su, Hao and Wang, Xiaolong and Chao, Yu-Wei and Fox, Dieter},
  booktitle={Robotics: Science and Systems (RSS)},
  year={2023}
}

@article{ryu2010_6dof_spacemouse,
  title={Development of a six {DOF} haptic master for teleoperation of a mobile manipulator},
  author={Ryu, Dongseok and Song, Jae-Bok and Cho, Changhyun and Kang, Sungchul and Kim, Munsang},
  journal={Mechatronics},
  year={2010},
  publisher={Elsevier}
}

@inproceedings{dhat2024using3dmice,
  title={Using {3D} mice to control robot manipulators},
  author={Dhat, Varad and Walker, Nick and Cakmak, Maya},
  booktitle={Proceedings of the 2024 ACM/IEEE International Conference on Human-Robot Interaction},
  year={2024}
}

@inproceedings{long2016effect_of_control_device,
  title={The effect of control device on performance in a robotic arm task},
  author={Long, Shelby K and Bliss, James P},
  booktitle={Proceedings of the Human Factors and Ergonomics Society Annual Meeting},
  year={2016},
  organization={SAGE Publications Sage CA: Los Angeles, CA}
}

@inproceedings{maccio2024kinesthetic,
  title={Kinesthetic Teaching in Robotics: a Mixed Reality Approach},
  author={Macci{\`o}, Simone and Shaaban, Mohamad and Carf{\`\i}, Alessandro and Mastrogiovanni, Fulvio},
  booktitle={2024 33rd IEEE International Conference on Robot and Human Interactive Communication (RO-MAN)},
  year={2024},
  organization={IEEE}
}

@inproceedings{yang2023moma_force,
  title={{MoMa-Force}: Visual-force imitation for real-world mobile manipulation},
  author={Yang, Taozheng and Jing, Ya and Wu, Hongtao and Xu, Jiafeng and Sima, Kuankuan and Chen, Guangzeng and Sima, Qie and Kong, Tao},
  booktitle={2023 IEEE/RSJ International Conference on Intelligent Robots and Systems (IROS)},
  year={2023},
  organization={IEEE}
}

@article{engelbracht2025spot_on,
  title={{Spot-On}: A Mixed Reality Interface for Multi-Robot Cooperation},
  author={Engelbracht, Tim and Lukovic, Petar and Behrens, Tjark and Lascheit, Kai and Zurbr{\"u}gg, Ren{\'e} and Pollefeys, Marc and Blum, Hermann and Bauer, Zuria},
  journal={arXiv preprint arXiv:2505.22539},
  year={2025}
}

@inproceedings{chen2023mr_teaming,
  title={A {3D} Mixed Reality Interface for Human-Robot Teaming},
  author={Chen, Jiaqi and Sun, Boyang and Pollefeys, Marc and Blum, Hermann},
  booktitle={2024 IEEE International Conference on Robotics and Automation (ICRA)},
  year={2024},
  organization={IEEE}
}

@inproceedings{gu2022ar_point_click,
  title={{AR} Point \& Click: An interface for setting robot navigation goals},
  author={Gu, Morris and Croft, Elizabeth and Cosgun, Akansel},
  booktitle={International Conference on Social Robotics},
  year={2022},
  organization={Springer}
}

@inproceedings{kemp2008point,
  title={A point-and-click interface for the real world: laser designation of objects for mobile manipulation},
  author={Kemp, Charles C and Anderson, Cressel D and Nguyen, Hai and Trevor, Alexander J and Xu, Zhe},
  booktitle={Proceedings of the 3rd ACM/IEEE international conference on human robot interaction},
  year={2008}
}

@article{lipton2017baxter,
  title={Baxter's homunculus: Virtual reality spaces for teleoperation in manufacturing},
  author={Lipton, Jeffrey I and Fay, Aidan J and Rus, Daniela},
  journal={IEEE Robotics and Automation Letters},
  year={2017},
  publisher={IEEE}
}

@inproceedings{laghi2018shared,
  title={Shared-autonomy control for intuitive bimanual tele-manipulation},
  author={Laghi, Marco and Maimeri, Michele and Marchand, Mathieu and Leparoux, Clara and Catalano, Manuel and Ajoudani, Arash and Bicchi, Antonio},
  booktitle={2018 IEEE-RAS 18th International Conference on Humanoid Robots (Humanoids)},
  year={2018},
  organization={IEEE}
}

@inproceedings{lewis2011two,
  title={Two hands are better than one: Assisting users with multi-robot manipulation tasks},
  author={Lewis, Bennie and Sukthankar, Gita},
  booktitle={2011 IEEE/RSJ International Conference on Intelligent Robots and Systems (IROS)},
  year={2011},
  organization={IEEE}
}

@article{roldan2017multi,
  title={Multi-robot interfaces and operator situational awareness: Study of the impact of immersion and prediction},
  author={Rold{\'a}n, Juan Jes{\'u}s and Pe{\~n}a-Tapia, Elena and Mart{\'\i}n-Barrio, Andr{\'e}s and Olivares-M{\'e}ndez, Miguel A and Del Cerro, Jaime and Barrientos, Antonio},
  journal={Sensors},
    year={2017},
  publisher={MDPI}
}

@inproceedings{wickens2015stom_task_switching_model,
  title={Visual attention allocation between robotic arm and environmental process control: Validating the {STOM} task switching model},
  author={Wickens, Christopher and Vieanne, Alex and Clegg, Benjamin and Sebok, Angelia and Janes, Jessica},
  booktitle={Proceedings of the Human Factors and Ergonomics Society Annual Meeting},
  year={2015},
  organization={SAGE Publications Sage CA: Los Angeles, CA}
}

@article{posadas2025beavr,
  title={{BEAVR}: Bimanual, Multi-Embodiment, Accessible, Virtual Reality Teleoperation System for Robots},
  author={Posadas-Nava, Alejandro and Carrasco, Alejandro and Linares, Richard},
  journal={arXiv preprint arXiv:2508.09606},
  year={2025}
}

@article{hirao2023body_extension_two_mobile_manip,
  title={Body extension by using two mobile manipulators},
  author={Hirao, Yusuke and Wan, Weiwei and Kanoulas, Dimitrios and Harada, Kensuke},
  journal={Cyborg and Bionic Systems},
  year={2023},
  publisher={AAAS}
}

@article{cowan2010magical,
  title={The magical mystery four: How is working memory capacity limited, and why?},
  author={Cowan, Nelson},
  journal={Current Directions in Psychological Science},
  year={2010},
  publisher={Sage Publications Sage CA: Los Angeles, CA}
}

@article{wickens2008multiple,
  title={Multiple resources and mental workload},
  author={Wickens, Christopher D},
  journal={Human Factors},
  year={2008},
  publisher={SAGE Publications Sage CA: Los Angeles, CA}
}

@article{ros2,
    author = {Steven Macenski and Tully Foote and Brian Gerkey and Chris Lalancette and William Woodall},
    title = {{Robot Operating System 2}: Design, architecture, and uses in the wild},
    journal = {Science Robotics},
    year = {2022},
    doi = {10.1126/scirobotics.abm6074}
}
}

\onecolumn
\appendix
% \subsection{Latency and Threads}
% one thread for camera
% two threads for depth completion
% main thread.
\subsection{Task Descriptions}
% Appendixes should appear before the acknowledgment.
\begin{table}[H]
    \centering
    \small
    \begin{tabular}{m{0.3cm}m{2.5cm}m{4cm}m{3cm}m{3cm}}
        \toprule
            \# & Name & Description & Initial Scene & Final Scene \\
        \midrule
            1 & Plush Toy Handoff & Initially, a white plush toy is placed in front of the robots. The operator needs to pick up the toy with one hand and hand it over to the other robot. & \includegraphics[width=\linewidth,valign=c]{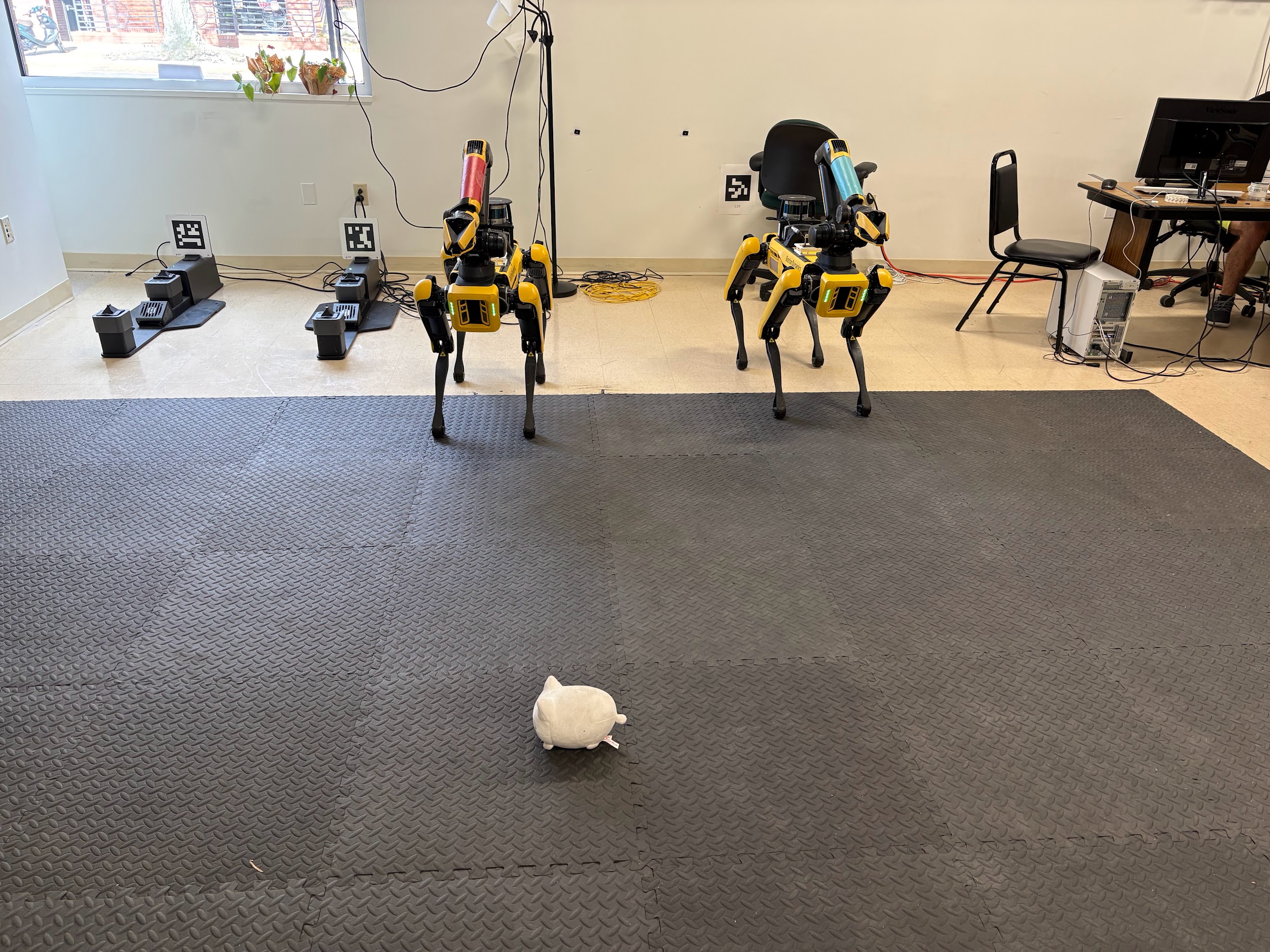} & \includegraphics[width=\linewidth,valign=c]{imgs/task_initial_end_imgs/1_final.jpg} \\
            \midrule
            2 & Large Box Picking and Placing & Initially, a large box is placed in front of the two robots. The box is too large that it has to be carried by two robots. The operators need to pick up the box, carry it near the platform covered with orange cloth, and place it on top of the platform. & \includegraphics[width=\linewidth,valign=c]{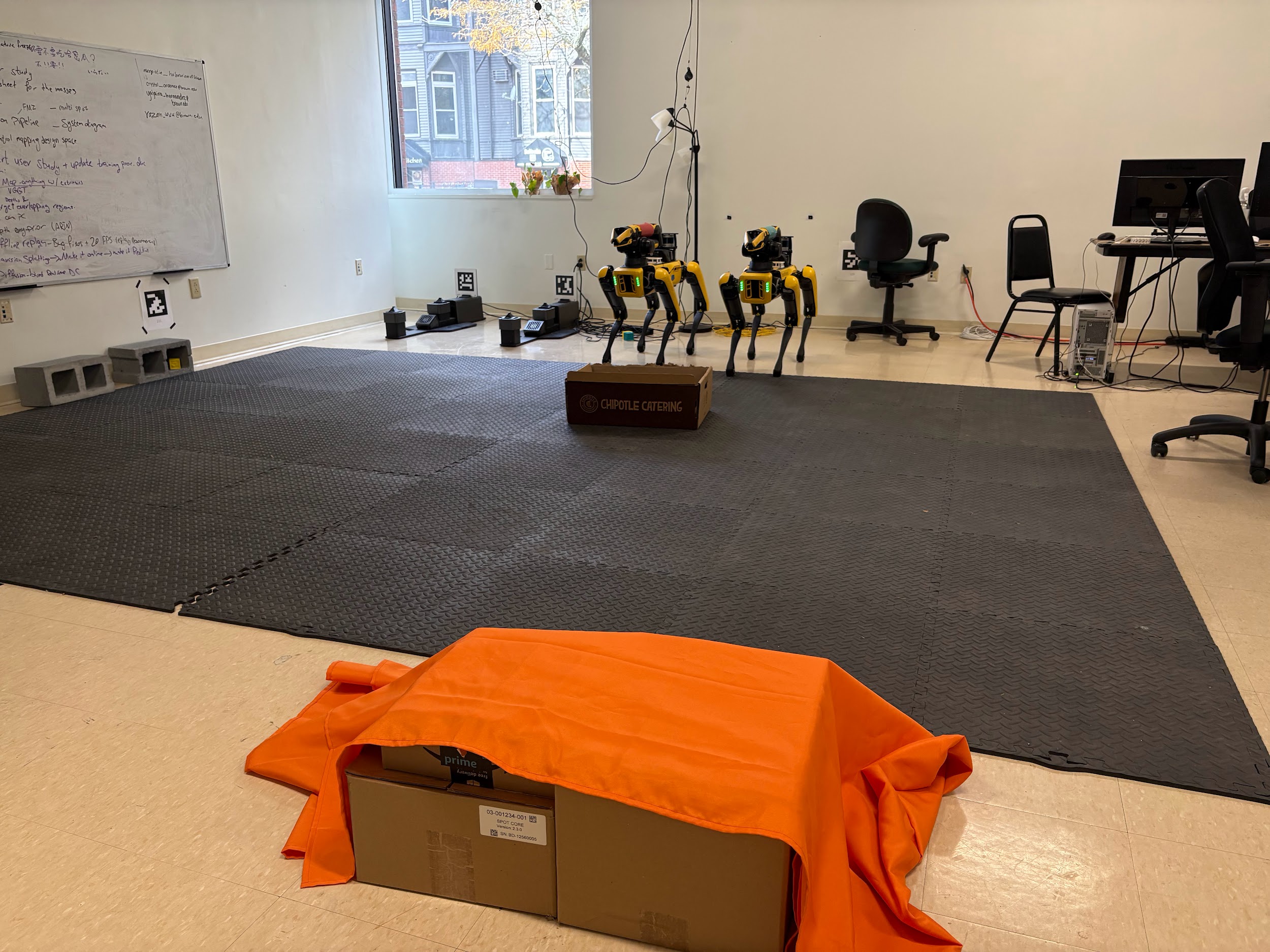} & \includegraphics[width=\linewidth,valign=c]{imgs/task_initial_end_imgs/2_final.jpg} \\
            \midrule
            3 & 3-Cube Stacking & Initially, three cubes are placed in a line in front of the robot. The operator needs to stack the three cubes in one stack in whatever order they deem best. & \includegraphics[width=\linewidth,valign=c]{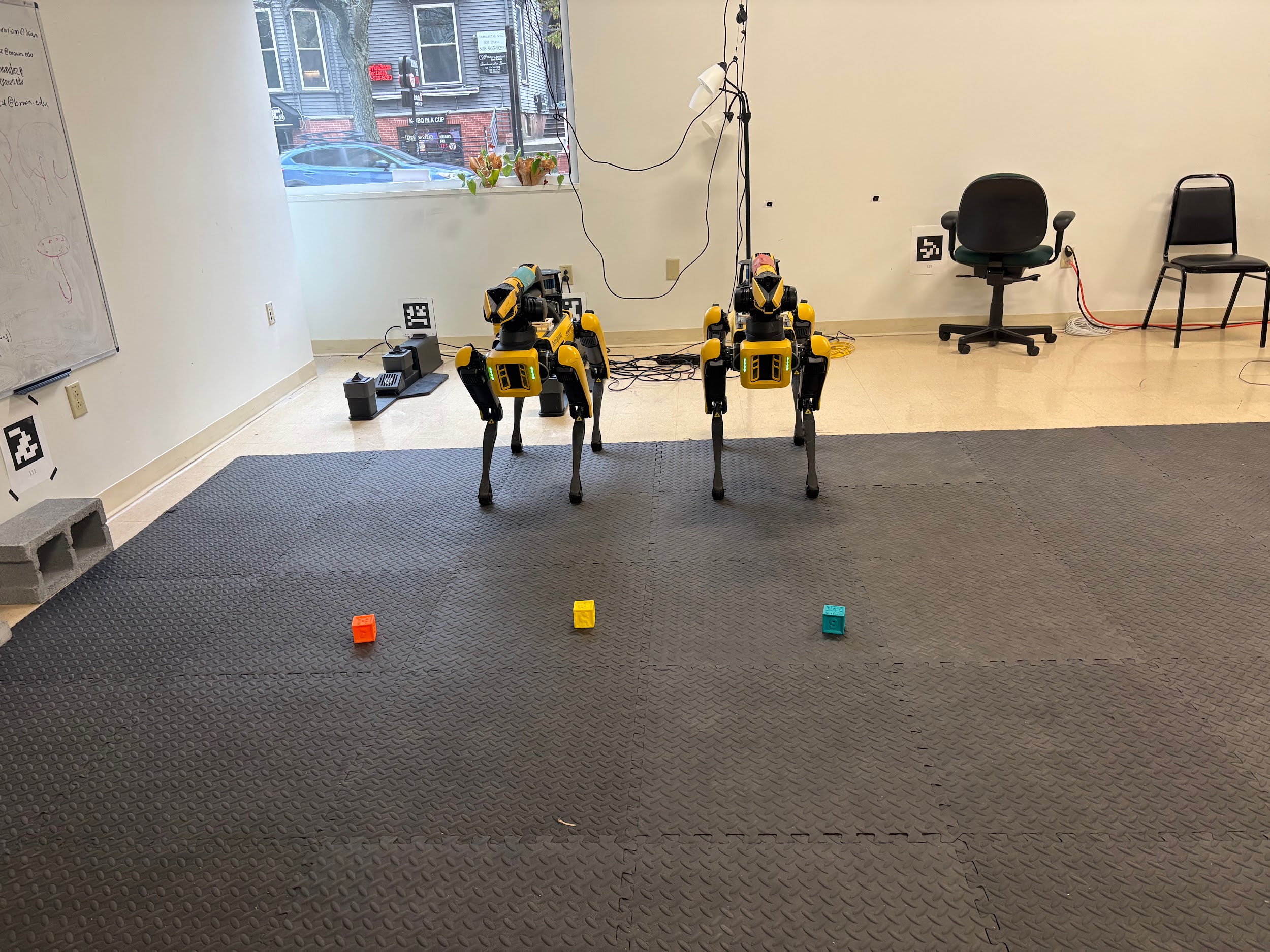} & \includegraphics[width=\linewidth,valign=c]{imgs/task_initial_end_imgs/4_final.jpg} \\
            \midrule
            4 & Collect 3 Scattered Cubes into the Basket & Initially, a basket and three cubes are scattered around in the scene in front of the robot. The operator is required to navigate and find all three cubes and place them into the basket in the center. & \includegraphics[width=\linewidth,valign=c]{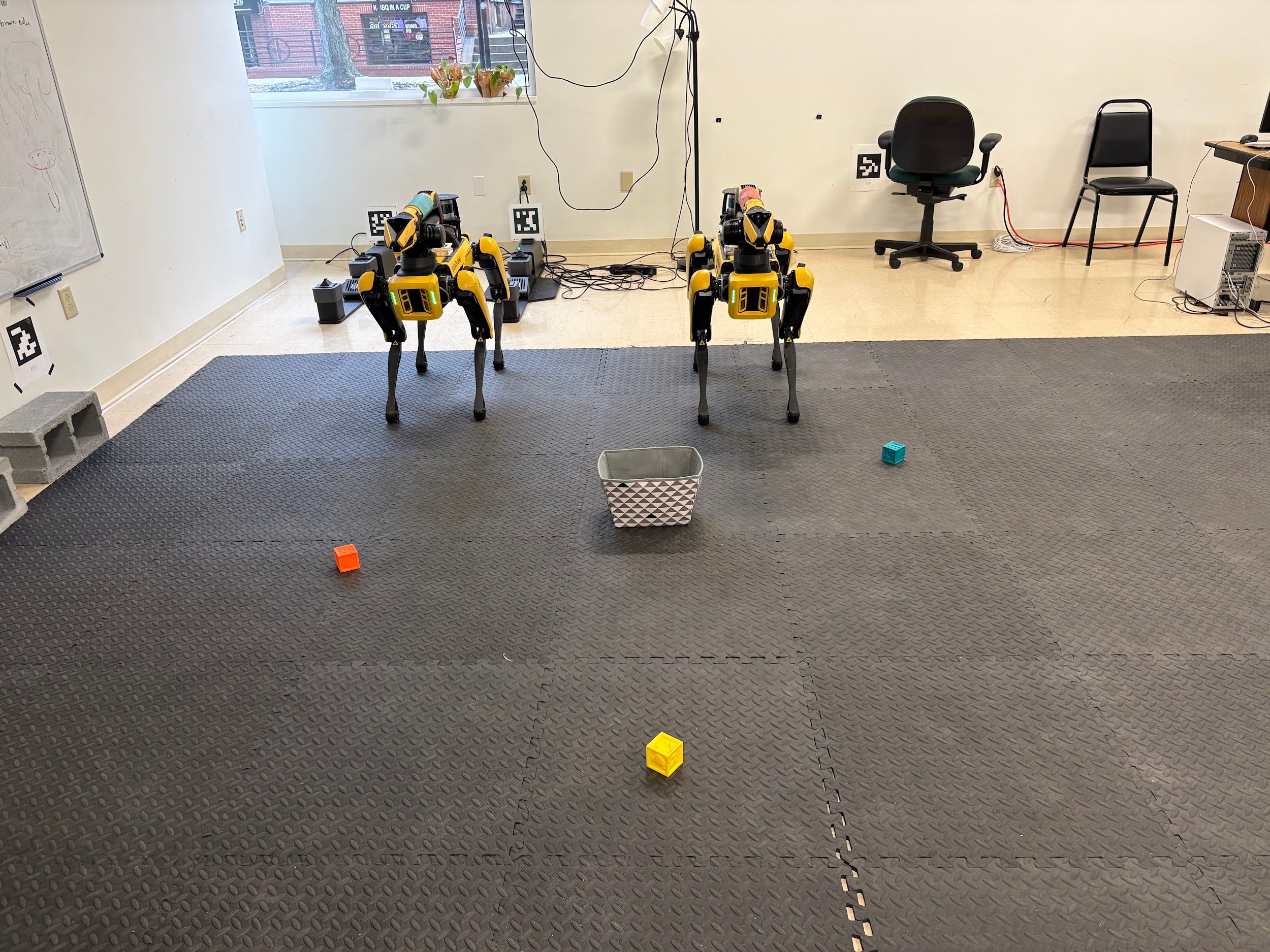} & \includegraphics[width=\linewidth,valign=c]{imgs/task_initial_end_imgs/3_final.jpg} \\
            \midrule
            5 & Opening and Holding Bags While Filling Them with 2 Cubes & Initially, one white plastic bag and two cubes are placed in front of the robots. The plastic bag is almost flattened, so it requires one robot to hold up the bag in order to have the opening big enough to fit the gripper and the cube. Thus, the operator will have to use one robot to hold up the bag, while the other robot looks around and searches for the cubes, and then places them into the held-up bag. & \includegraphics[width=\linewidth,valign=c]{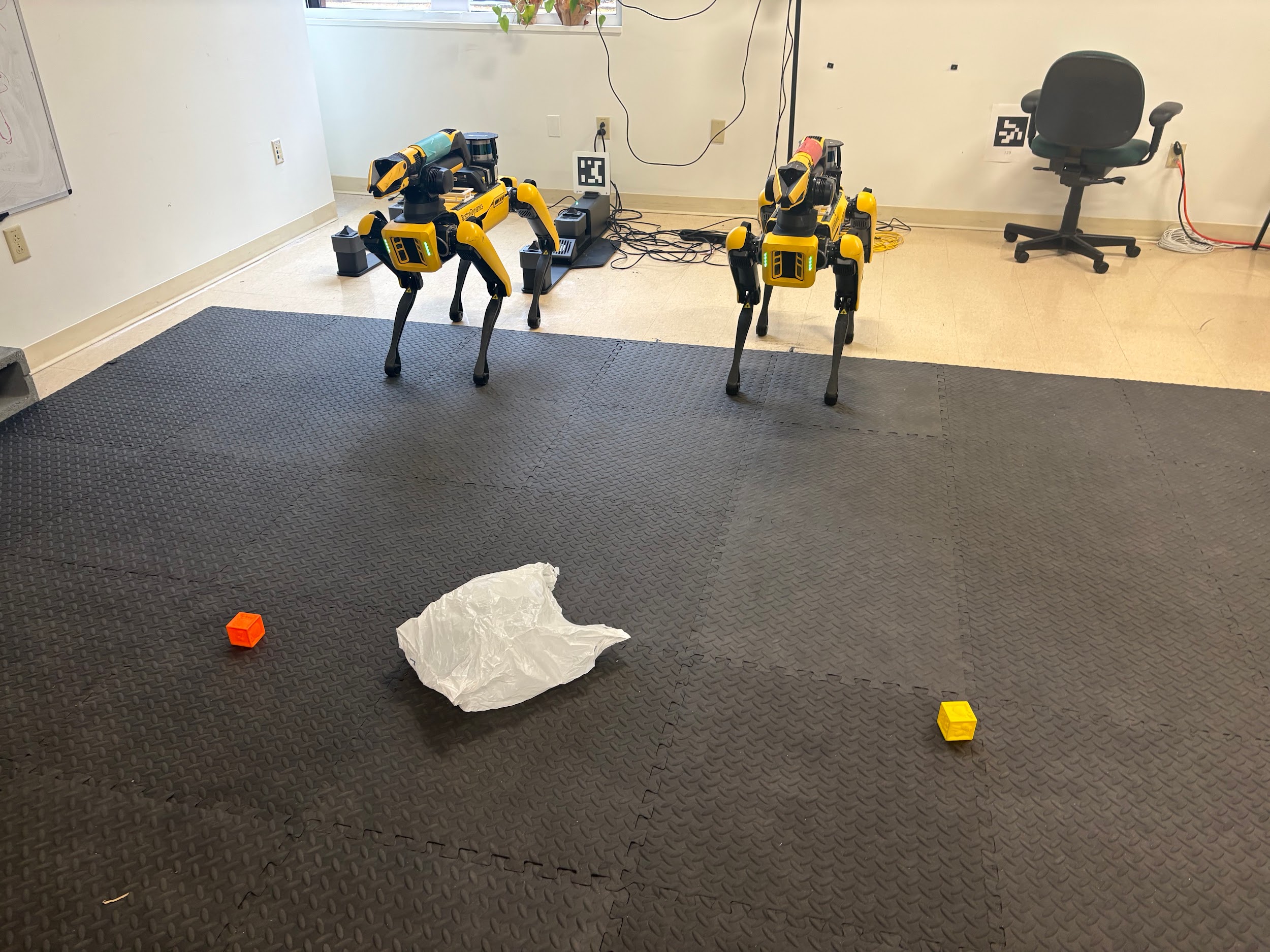} & \includegraphics[width=\linewidth,valign=c]{imgs/task_initial_end_imgs/6_final.jpg} \\
            \midrule
            6 & Weighted Door Opening and Pass Through & The robots need to open a weighted door and make one robot go through the door. The door is too heavy for the Boston Dynamics Spot's automatic door-opening function, which will not properly open the door and let the robot through. & \includegraphics[width=\linewidth,valign=c]{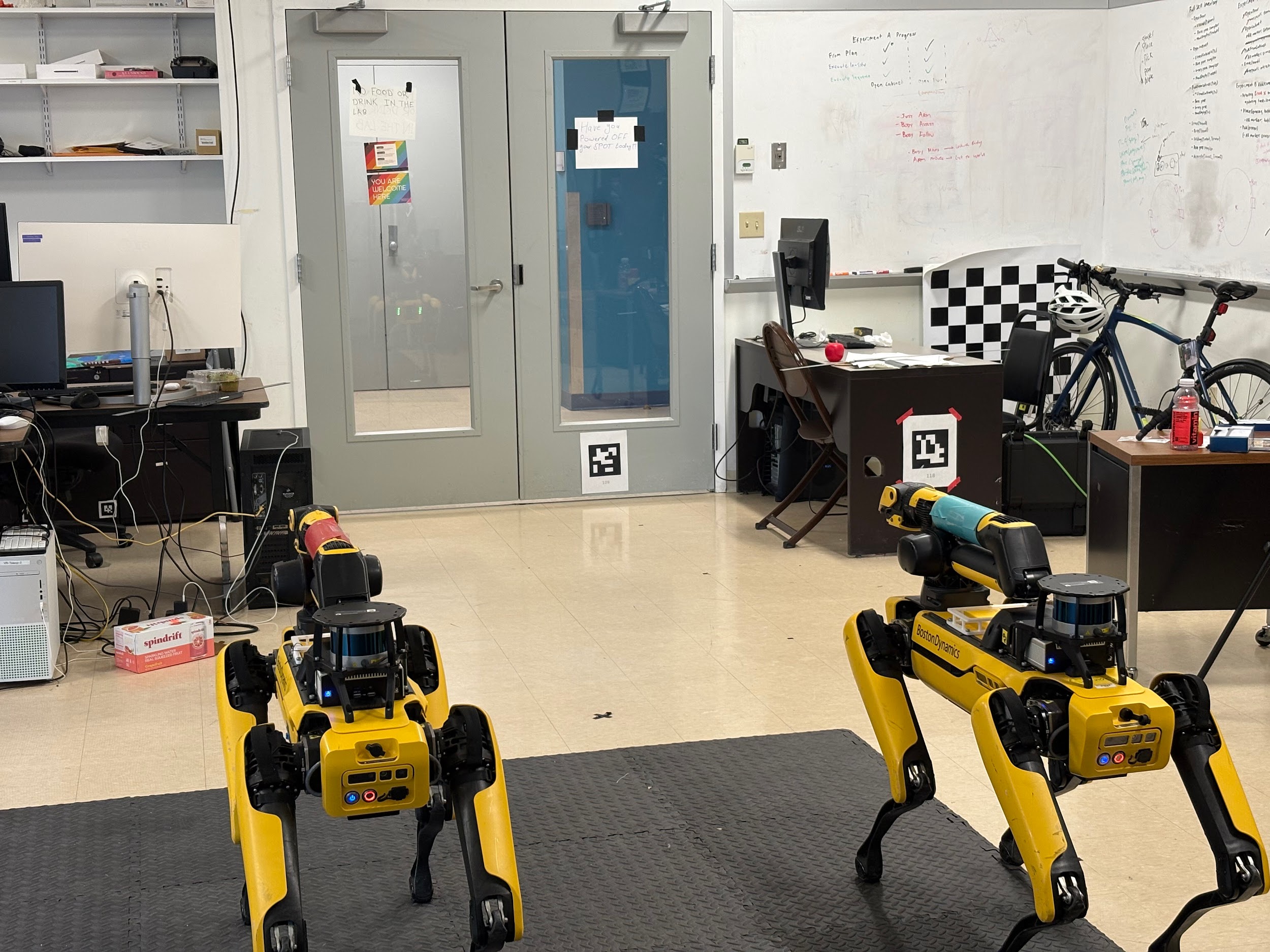} & \includegraphics[width=\linewidth,valign=c]{imgs/task_initial_end_imgs/7_final.jpg} \\
        \bottomrule
    \end{tabular}
    \caption{Task Descriptions for each task (Part 1)}
\end{table}

\begin{table}[H]
    \centering
    \small
    \begin{tabular}{m{0.3cm}m{2.5cm}m{4cm}m{3cm}m{3cm}}
        \toprule
            \# & Name & Description & Initial Scene & Final Scene \\
            \midrule
            7 & Dust Pan Sweep & Initially, two cubes, one broom, and one dust pan are placed in front of the robot. The operator needs to control the robot to use the broom to sweep the two cubes into the dust pan. & \includegraphics[width=\linewidth,valign=c]{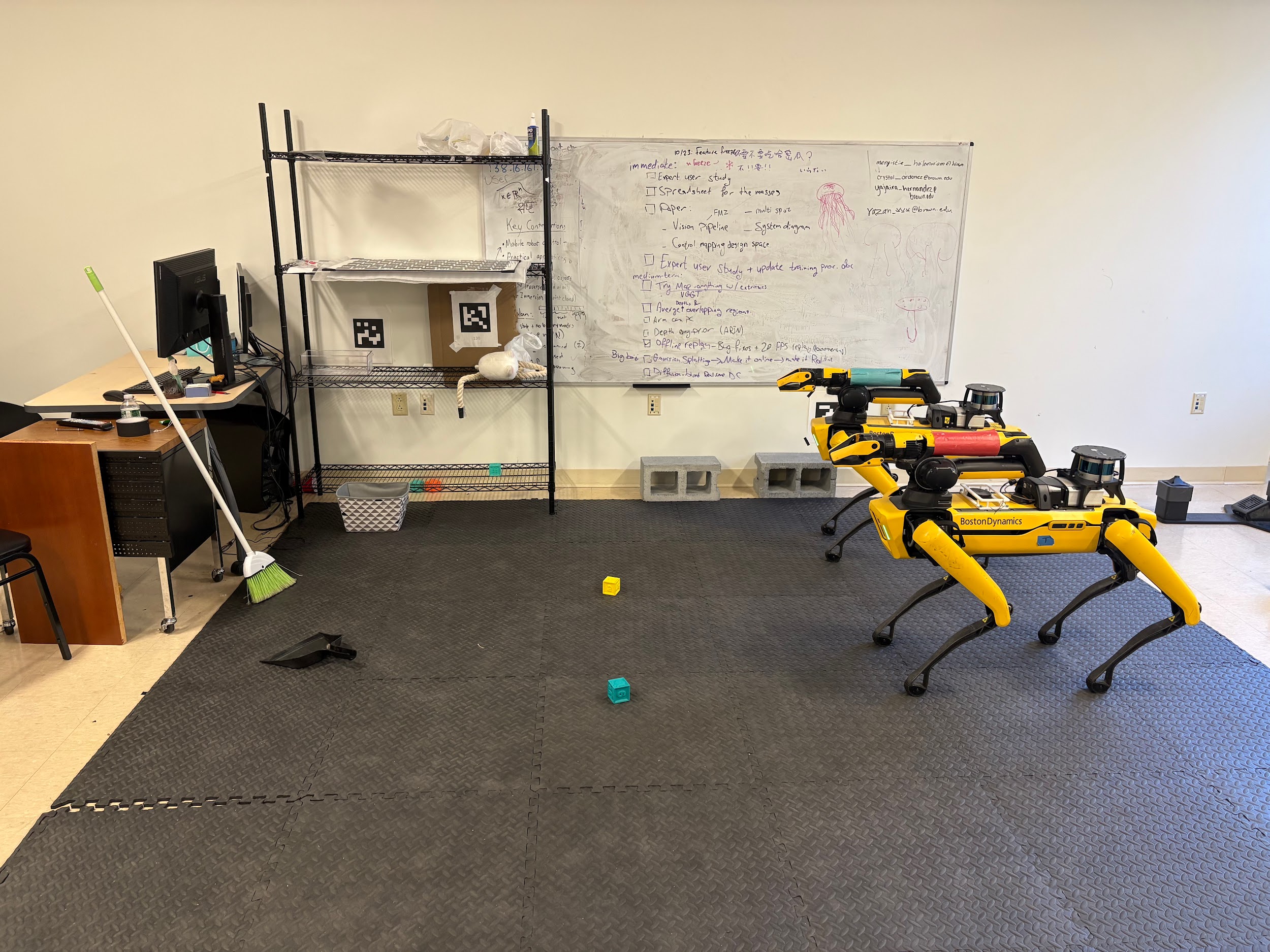} & \includegraphics[width=\linewidth,valign=c]{imgs/task_initial_end_imgs/8_final.jpg} \\
            \midrule
            8 & Bedsheet Folding & Initially, a very large, long sheet is placed in front of the robots. The operator needs to fold it in half. Note that since the sheet is so large, the robot will have to move the base in order to properly fold the sheet. & \includegraphics[width=\linewidth,valign=c]{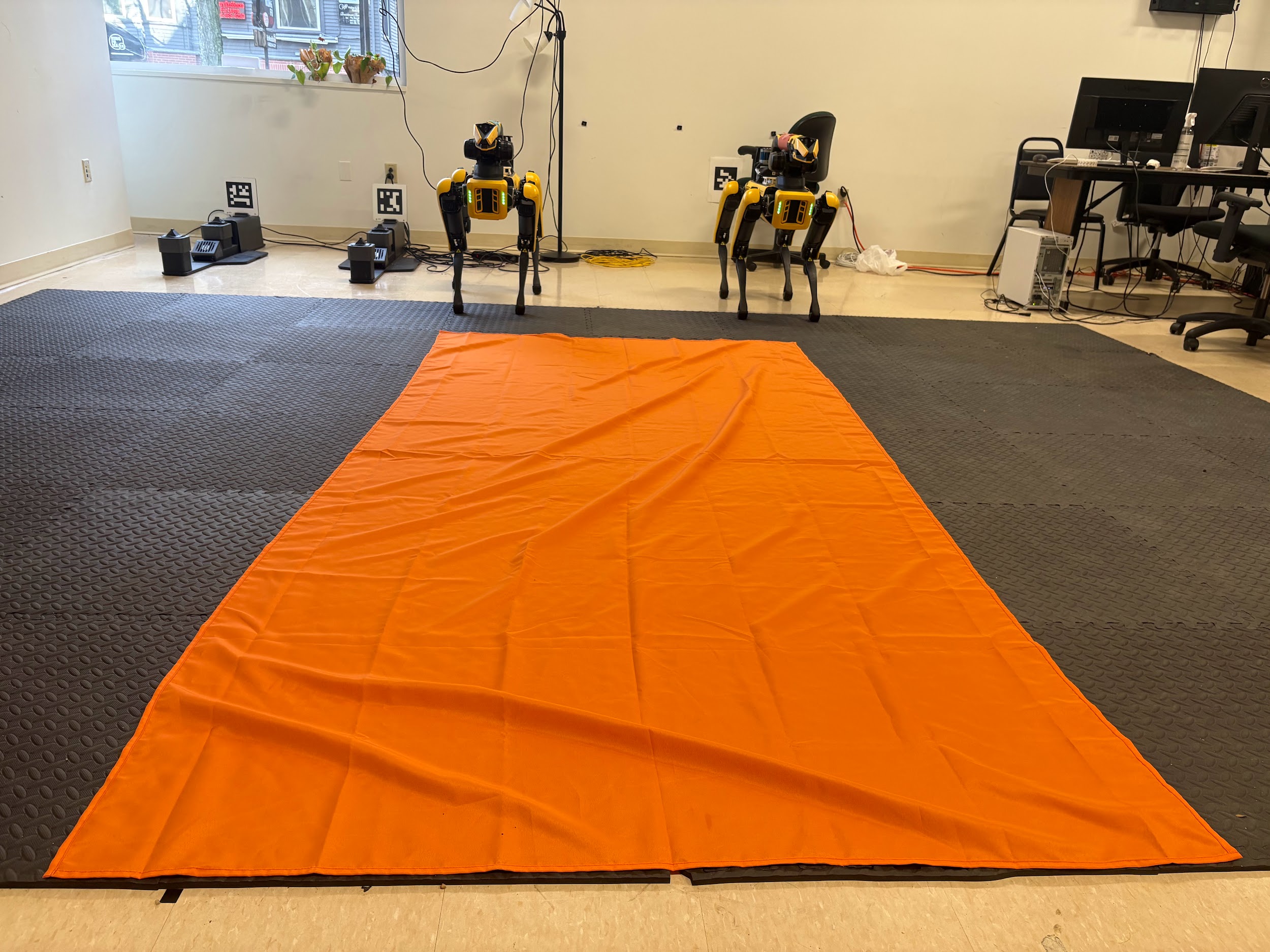} & \includegraphics[width=\linewidth,valign=c]{imgs/task_initial_end_imgs/5_final.jpg} \\
            \midrule
            9 & Rope Tying 1 Knot & Initially, a thick rope is placed in front of the two robots. The operator needs to control the robot and tie a knot on the rope. & \includegraphics[width=\linewidth,valign=c]{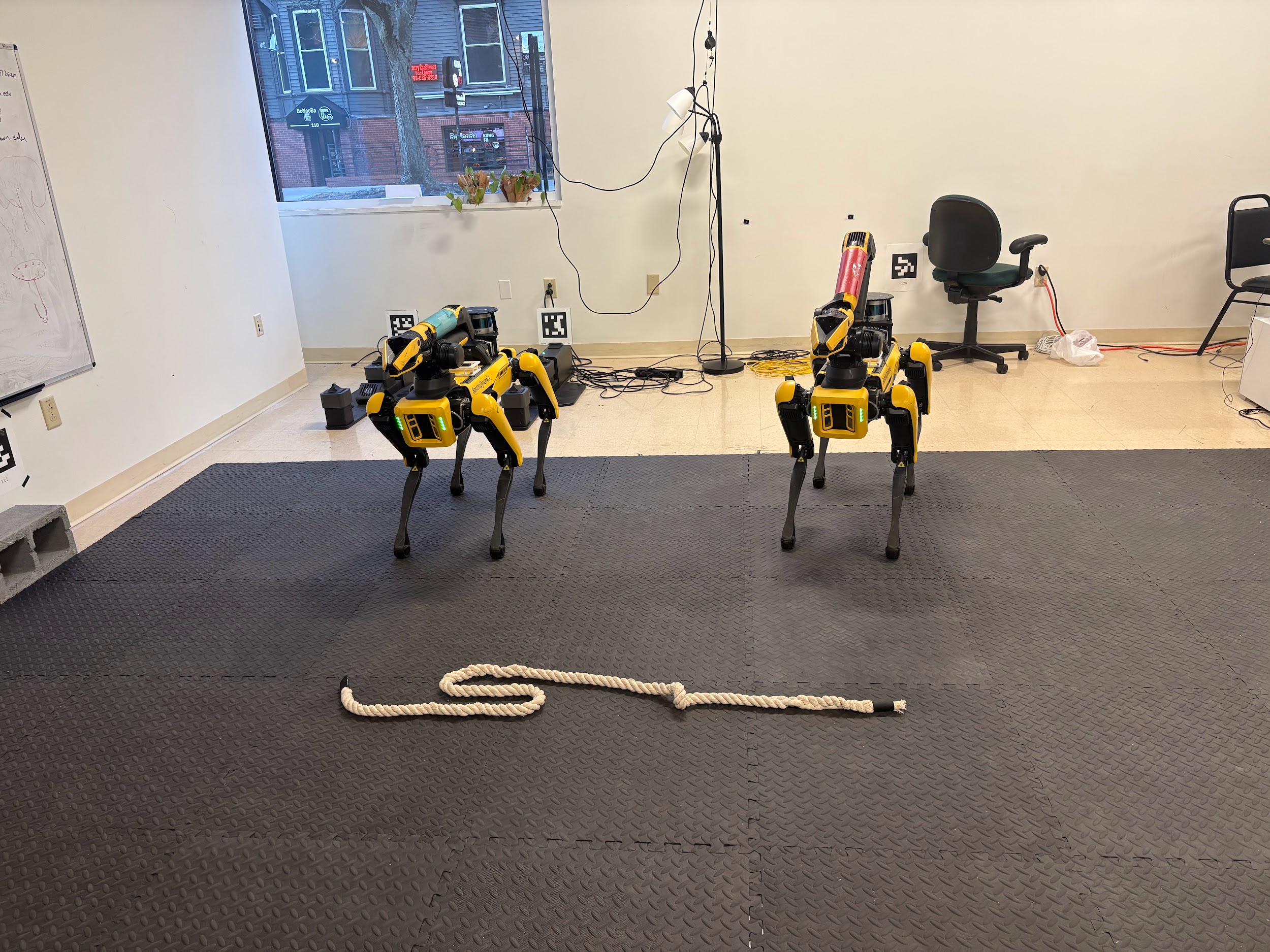} & \includegraphics[width=\linewidth,valign=c]{imgs/task_initial_end_imgs/9_final.jpg} \\
        \bottomrule
    \end{tabular}
    \caption{Task Descriptions for each task (Part 2)}
\end{table}

%%%%%%%
\end{document}